\pdfoutput=1
\documentclass[lettersize,journal]{IEEEtran}
\usepackage{amsmath,amsfonts,amsthm}
\usepackage{amssymb}
\usepackage{stmaryrd}
\usepackage{array}
\usepackage{textcomp}
\usepackage{stfloats}
\usepackage{url}
\usepackage{verbatim}
\usepackage{graphicx}
\usepackage{multirow}
\usepackage{threeparttable}
\usepackage{xcolor} 
\usepackage{tikz}
\usepackage{cite}
\renewcommand{\baselinestretch}{0.973}

\usepackage{subcaption}
\usepackage{algorithm}
\usepackage{algpseudocode} 
\usepackage{xspace}

\usepackage[acronym]{glossaries-extra}
\setabbreviationstyle[acronym]{long-short}
\glssetcategoryattribute{acronym}{nohyperfirst}{true}

\usepackage{hyperref}                          
\usepackage[capitalise, noabbrev]{cleveref}    
\makeatletter
    \if@cref@capitalise
        \crefname{subsection}{Section}{Sections}
        \crefname{subsubsection}{Section}{Sections}
        \crefname{assumption}{Assumption}{Assumptions}
        \crefname{problem}{Problem}{Problems}
    \else
        \crefname{subsection}{section}{sections}
        \crefname{subsubsection}{section}{sections}
        \crefname{assumption}{assumption}{assumptions}
        \crefname{problem}{problem}{problems}
    \fi
\makeatother
\theoremstyle{definition}
\newtheorem{definition}{\textbf{Definition}}
\newtheorem{remark}{\textbf{Remark}}

\newtheorem{theorem}{\textbf{Theorem}}
\usepackage{mdframed}
\usepackage{booktabs}
\usepackage{siunitx}
\usepackage{tabularx}
\definecolor{green}{HTML}{008000}
\definecolor{blue}{HTML}{3963c8}
\definecolor{lightblue}{HTML}{c1ddf5}
\definecolor{yellow}{HTML}{ffa500}
\definecolor{orange}{HTML}{FF9410}

\definecolor{revised}{rgb}{0.0, 0.0, 1.0}
\newcommand{\rebuttal}[1]{{\color{revised}{}#1}}
\newenvironment{rebuttalblock}{\color{revised}}{}
\makeatletter
\newcommand{\hideoriginaltext}{\DeclareRobustCommand{\ori}[1]{\@bsphack\@esphack}}
\newcommand{\hiderevisiontags}{\DeclareRobustCommand{\revtag}[1]{\@bsphack\@esphack}}
\makeatother
\newcommand{\hiderevisionmarks}{%
    \renewcommand{\rebuttal}[1]{##1}%
    \hiderevisiontags
    \hideoriginaltext
}
\newcommand{\revtag}[1]{\textcolor{red}{\,\mbox{[#1]}}}

\newcommand{\resource}[1]{\textcolor{red}{#1}}
\newcommand{\functionality}[1]{{\textcolor{green}{#1}}}
\newcommand{\implementation}[1]{{\textcolor{orange}{#1}}}

\newcommand{\tab}{\hspace{0.2cm}}
\newcommand{\tabn}[1]{\ifnum#1>0 \tab\expandafter\tabn\expandafter{\number\numexpr#1-1}\fi}

\newcommand{\maketup}[1]{\langle #1 \rangle}
\newcommand{\makeset}[1]{\left\{ #1 \right\}}

\newcommand{\DP}{\mathcal{DP}}

\newcommand{\dprb}{{dp}}

\newcommand{\DistOf}{\Delta}
\newcommand{\Prob}{\mathbb{P}}
\newcommand{\outcomeDist}{\Prob}
\newcommand{\inducedDist}[2]{\Prob\left(#1 \;\middle|\; #2\right)}

\newcommand{\trainsetting}{s}
\newcommand{\trainsettingSet}{S}

\newcommand{\chipconfig}{h}
\newcommand{\chipconfigSet}{H}

\newcommand{\fabconfig}{f}
\newcommand{\fabconfigSet}{F}

\DeclareRobustCommand{\bluecircled}[1]{%
    \tikz[baseline=(char.base)]{%
        \node[
            shape=circle,
            fill=lightblue,
            inner sep=1pt
        ] (char) {#1};%
    }%
}

\definecolor{dpred}{rgb}{0.7, 0.0, 0.0}
\definecolor{dpgreen}{rgb}{0.0, 0.5, 0.0}
\definecolor{functorpurple}{rgb}{0.5, 0, 0.5}
\definecolor{imporange}{rgb}{1.0, 0.58, 0.063}
\definecolor{specificationcolor}{rgb}{0.258, 0.527, 0.957}

\newcommand{\op}{\text{op}}
\newcommand{\defeq}{\mathrel{\raisebox{-0.3ex}{$\overset{\text{\tiny def}}{=}$}}}

\newcommand{\setWithArg}[2]{\{#1 \mid #2\}}
\newcommand{\setWithIndex}[2]{\{#1\}_{#2}}

\newcommand{\posetP}{P}
\newcommand{\posetQ}{Q}
\newcommand{\posetR}{R}

\newcommand{\posetleq}{\preceq}

\newcommand{\subsetX}{X}

\newcommand{\subsetXP}{\subsetX_\posetP}

\newcommand{\poselxP}{x_{\posetP}}
\newcommand{\poselyP}{y_{\posetP}}
\newcommand{\poselxQ}{x_{\posetQ}}
\newcommand{\poselyQ}{y_{\posetQ}}
\newcommand{\poselxRnoRes}{x_{\posetR}}

\newcommand{\upperClosure}[1]{\uparrow\!#1}

\newcommand{\dpOf}[2]{\DP\{#1, #2\}}

\newcommand{\dprbOf}[1]{\dprb_#1}
\newcommand{\dprba}{\dprbOf{a}}
\newcommand{\dprbb}{\dprbOf{b}}

\newcommand{\union}{\vee}
\newcommand{\intersection}{\wedge}

\newcommand{\mthen}{\fatsemi}
\newcommand{\mthenMathbin}{\mathbin{\mthen}}
\newcommand{\stack}{\otimes}
\newcommand{\stackMathbin}{\mathbin{\stack}}
\newcommand{\trace}{\text{Tr}}
\newcommand{\traceOf}[1]{\trace(#1)}

\usepackage[normalem]{ulem}
\definecolor{oritext}{cmyk}{0.62,0.57,0.23,0}   %
\makeatletter
\newcommand{\ori@box}[1]{%
    \@ifundefined{\expandafter\@gobble\string#1}{}{%
        \expandafter\DeclareCommandCopy\csname ori@saved\string#1\endcsname#1%
        \edef\ori@tmp{\expandafter\noexpand\csname ori@saved\string#1\endcsname}%
        \expandafter\ori@redefine\expandafter{\ori@tmp}#1}%
}
\newcommand{\ori@redefine}[2]{%
    \RenewDocumentCommand#2{s o m}{\mbox{\IfBooleanTF{##1}%
        {\IfValueTF{##2}{#1*[{##2}]{##3}}{#1*{##3}}}%
        {\IfValueTF{##2}{#1[{##2}]{##3}}{#1{##3}}}}}%
}
\newcommand{\ori@protect}{%
    \ori@box\cite\ori@box\footnote\ori@box\ref\ori@box\pageref
    \ori@box\eqref\ori@box\cref\ori@box\Cref
}
\DeclareRobustCommand{\ori}[1]{{\color{oritext}\ori@protect\sout{#1}}}
\makeatother

\hideoriginaltext
\hiderevisiontags
\hiderevisionmarks

\newcommand{\newacronymwithcmds}[3]{%
  \newacronym{abk:#1}{#2}{#3}%
  \expandafter\newcommand\csname gls#1\endcsname{\gls{abk:#1}\xspace}
  \expandafter\newcommand\csname Gls#1\endcsname{\Gls{abk:#1}\xspace}
  \expandafter\newcommand\csname glspl#1\endcsname{\glspl{abk:#1}\xspace}
  \expandafter\newcommand\csname Glspl#1\endcsname{\Glspl{abk:#1}\xspace}
}

\newcommand{\glsdefhere}[2]{%
  \glsreset{abk:#2}%
  \hypertarget{glo:abk:#2}{}%
  #1{abk:#2}%
  \glsunset{abk:#2}%
}

\newacronymwithcmds{poset}{poset}{partially ordered set}

\newacronymwithcmds{dp}{DP}{design problem}

\newacronymwithcmds{mdpi}{MDPI}{monotone design problem with implementation}

\newacronymwithcmds{sgd}{SGD}{stochastic gradient descent}

\newacronymwithcmds{ga}{GA}{genetic algorithm}

\newacronymwithcmds{nnp}{NNP}{neural network processor}

\newacronymwithcmds{lea}{LEA}{latency, energy consumption, and area}

\newacronymwithcmds{pdf}{PDF}{probability density function}

\newacronymwithcmds{cdf}{CDF}{cumulative distribution function}

\newacronymwithcmds{rmse}{RMSE}{root mean square deviation}

\newacronymwithcmds{pe}{PE}{processing element}

\newacronymwithcmds{lone}{L1}{local buffer}

\newacronymwithcmds{ltwo}{L2}{global buffer}

\newacronymwithcmds{noc}{NoC}{network-on-chip}

\usetikzlibrary{positioning,intersections, hobby, patterns, calc, decorations.pathmorphing, decorations.markings, shadows,shapes, cd,arrows.meta, fit,quotes}

\tikzset{
   tick/.style={postaction={
      decorate,
      decoration={markings, mark=at position 0.5 with {\draw[-] (0,.4ex) -- (0,-.4ex);}}}
   }
}
\tikzstyle{block} = [draw, rectangle, minimum height=2em, minimum width=3em,font=\bfseries,rounded corners,thick]
\tikzstyle{block} = [draw, rectangle, minimum height=2em, minimum width=3em]
\tikzstyle{block1} = [draw, rectangle, minimum height=1.5em, minimum width=2.5em]
\tikzstyle{blockDyn} = [draw, rectangle, minimum height=2.5em, minimum width=3.5em, align=center, inner sep=10pt, thick, fill=white, copy shadow={draw=black,fill=black,opacity=1,shadow xshift=0.5ex,shadow yshift=-0.5ex}]
\tikzstyle{blockAlg} = [draw, rectangle, minimum height=1.5em, minimum width=2.5em, align=center, inner sep=10pt, thick]
\tikzstyle{sum} = [draw,circle]

\tikzstyle{nodePre} = [circle, draw,inner sep=1pt,node contents={$\preceq$},thick]
\tikzstyle{nodePreEmpty} = [circle, draw,inner sep=1pt,thick]
\tikzstyle{nodePos} = [circle, draw,inner sep=1pt,node contents={$\posceq$},thick]
\tikzstyle{nodeProd} = [rectangle, draw,inner sep=4pt,node contents={$\times$},rounded corners,thick]
\tikzstyle{nodeSum} = [rectangle, draw,inner sep=4pt,node contents={$\mathbf{+}$},rounded corners,thick]

\definecolor{red}{rgb}{0.75, 0.0, 0.0}

\tikzset{fcname/.store in =\fcname, fcname={}}
\tikzset{funame/.store in =\funame, funame={}}
\tikzset{rcname/.store in =\rcname, rcname={}}
\tikzset{runame/.store in =\runame, runame={}}
\tikzset{whereres/.store in =\whereres, whereres=0.5}
\tikzset{wherefun/.store in =\wherefun, wherefun=0.5}
\tikzset{relres/.store in =\relres, relres={above}}
\tikzset{relfun/.store in =\relfun, relfun={above}}
\tikzset{posres/.store in =\posres, posres=1}
\tikzset{posfun/.store in =\posfun, posfun=1}
\tikzset{loos/.store in =\loos, loos=2}
\tikzset{feedback/.store in =\feedback, feedback=0}
\tikzset{
   DP/.style={%
      label/.style={
         font=\everymath\expandafter{\the\everymath\scriptstyle},
         inner sep=5pt,
         node distance=2pt and -2pt},
      semithick,
      node distance=1 and 1,
      rconn/.style={color=white,opacity=0.0,postaction={decorate}, shorten <=3.2pt, shorten >= 0.8,
      decoration={markings, 
      mark= at position 0 with {
               \coordinate (a);
      },
      mark=at position .5 with
      {
              \ifthenelse{\equal{\feedback}{1}}{\def\angleOut{90}\def\angleIn{90}}{\def\angleOut{0}\def\angleIn{180}}    
              \coordinate (b);
              \draw[dashed,dpred,opacity=1.0] (a) to[out=\angleOut,in=\angleIn,looseness=\loos] 
              node[pos=\posres,\relres=\whereres mm,dpred,opacity=1,fill=white,inner sep=1pt,outer sep=1pt]{\footnotesize{\rcname}} (b);
      },
      mark= at position 1 with 
      {
             \ifthenelse{\equal{\feedback}{1}}{\def\angleOut{0}\def\angleIn{0}}{\def\angleOut{180}\def\angleIn{0}} 
              \ifthenelse{\equal{\feedback}{1}}{\def\symbol{\succeq}}{\def\symbol{\preceq}} 
              \coordinate (c);
              \draw[dpgreen,opacity=1.0] (c) to[out=\angleOut,in=\angleIn,looseness=\loos]
              node[pos=\posfun,\relfun=\wherefun mm,dpgreen,opacity=1,fill=white,inner sep=1pt,outer sep=1pt]{\footnotesize{\fcname}} (b){}; %
              \node[draw,circle,inner sep=0.5pt,color=black,fill=white,opacity=1.0] at (b) (nodepreceq) {$\symbol$}; 
      }
      }},
      runconn/.style={color=dpred,dashed,postaction={decorate},
      decoration={markings,
      mark= at position 1 with {
              \coordinate (a);
              \draw[dpred,opacity=1.0,dashed] ($(a)+(0.05,0)$) --++ (0.5,0) node[\relres,pos=\posres]{\footnotesize{\runame}};}
      }
      },
      funconn/.style={color=white,postaction={decorate},
      decoration={markings,
      mark= at position 0 with {
      \coordinate (a);
      \draw[dpgreen] ($(a)+(-0.05,0)$) -- ($(a)+(-0.5,0)$) node[\relfun, pos=\posfun]{\footnotesize{\funame}};}
      }
      },
      execute at begin picture={\tikzset{
         x=\dpx, y=\dpy,
         every fit/.style={inner xsep=\dpx, inner ysep=\dpy}}}
      },
   dpx/.store in=\dpx,
   dpx = 1.5cm,
   dpy/.store in=\dpy,
   dpy = 1.5ex,
   dp port sep/.store in=\dpportsep,
   dp port sep=2,
   dp port length/.store in=\dpportlen,
   dp port length=4pt,
   dp min width/.store in=\dpminwidth,
   dp min width=0.5cm,
   dp rounded corners/.store in=\dpcorners,
   dp rounded corners=2pt,
   dp small/.style={dp port sep=1, dp port length=2.5pt, dpx=.4cm, dp min width=.4cm, dpy=.7ex},
   dp/.code 2 args={%
      \pgfmathsetlengthmacro{\dpheight}{\dpportsep * (max(#1,#2)) * \dpy}
      \pgfkeysalso{draw,%
        minimum width=\dpminwidth,%
        minimum height=\dpheight,%
        font=\bfseries,
        outer sep=0pt,%
        inner sep=5pt,%
        rounded corners=\dpcorners,
        thick,
        prefix after command={\pgfextra{\let\fixname\tikzlastnode}},
        append after command={\pgfextra{\draw
            \ifnum #1=0{} \else foreach \i in {1,...,#1} { 
            ($(\fixname.north west)!{\i/(#1+1)}!(\fixname.south west)$) +(0,0) node[solid,left,circle,color=dpgreen,draw,fill=dpgreen,scale=0.3] {} coordinate (\fixname_fun\i) -- +(0,0) coordinate (\fixname_fun\i')}\fi %
            \ifnum #2=0{} \else foreach \i in {1,...,#2} {
            ($(\fixname.north east)!{\i/(#2+1)}!(\fixname.south east)$) +(0,0) coordinate (\fixname_res\i') -- +(0,0) node[solid,right,circle,color=dpred,draw,fill=dpred,scale=0.3] {} coordinate (\fixname_res\i)}\fi;
         }}}
         },
      dp name/.style={append after command={\pgfextra{\node[label=center,inner sep=2pt,fill=white] at (\fixname) {\textbf{#1}};}}}
   }

\begin{document}

\title{Uncertainty-Aware End-to-End Co-Design of Neural Network Processors: From Training and Mapping to Fabrication}

\author{Yuyang Du$^1$, Yujun Huang$^2$, Gioele Zardini$^2$
\thanks{$^1$School of Electrical and Electronic Engineering, Nanyang Technological University, Singapore, {\tt yuyang006@e.ntu.edu.sg}}
\thanks{$^2$Laboratory for Information \& Decision Systems,
Massachusetts Institute of Technology, Cambridge, MA 02139 {\tt \{yujun233,gzardini\}@mit.edu}}
\thanks{This material is based on research supported by the Defense Advanced Research Projects Agency (DARPA) under Award D25AC00373.}
}

\maketitle

\begin{abstract}
Designing a neural network processor is an end-to-end co-design problem:  network architecture and training budget determine the inference workload;  hardware mapping decisions determine chip area, latency, and energy; and  these characteristics govern fabrication yield and manufacturing cost.
In practice, these decisions are made in separate stages, and existing  co-design methodologies are tightly coupled to specific algorithms, making  it difficult to improve one component without reworking the entire pipeline.
This paper presents a unified framework, grounded in monotone co-design  theory, that composes five interoperable design blocks spanning network  training, chip mapping, wafer-level fabrication, migration, and compute resource  allocation. 
Each block exposes only a functionality-resource interface to  the rest of the system, so any block can be refined without structural  changes elsewhere. 
A central contribution is the treatment of uncertainty:  rather than collapsing stochastic outcomes into point estimates, the framework introduces Confidence, the inverse of success probability, as an  explicit and optimizable resource alongside cost, time, and power.
Three case studies validate the approach. 
The first recovers Pareto-optimal  implementations across heterogeneous application scenarios. 
The second  confirms that Confidence functions as a continuously tunable design knob  rather than a post-hoc diagnostic. 
The third demonstrates that improving a  single block's implementation set automatically propagates to the global 
Pareto front, without modifying the co-design diagram.
\end{abstract}


\section{Introduction}\label{sec:intro}
\IEEEPARstart{D}{esigning} a \glsnnp is inherently an end-to-end co-design problem. 
The selected neural network and its training budget determine the workload that must be supported at deployment; hardware mapping decisions determine chip area, latency, and energy; and these hardware characteristics, in turn, affect fabrication yield and manufacturing cost. 
In practice, however, these decisions are still often taken in separate stages and with limited feedback across teams. 
This separation is increasingly problematic as application-specific inference platforms replace general-purpose computation platforms like GPUs in cost-, power-, and latency-sensitive domains. 
What is needed is a principled way to compose decisions from network training to hardware mapping and fabrication.

\begin{figure}[t] 
    \centering\includegraphics[width=\columnwidth]{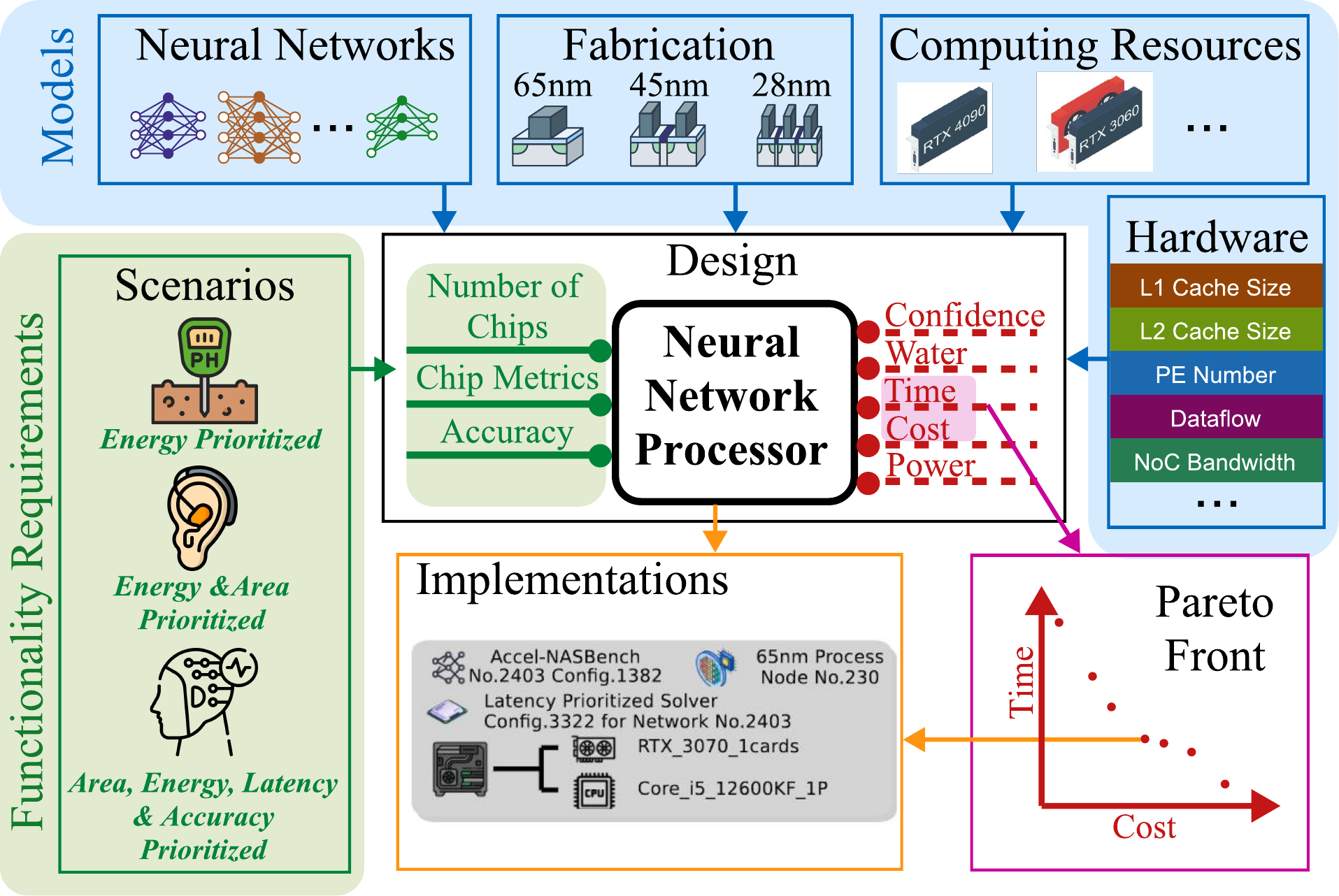} 
    \caption{Graphical illustration of the informal problem definition for co-designing a \glsnnp. Application scenarios (left) specify the functionality requirements for this design; by modeling neural networks, hardware, fabrication, and computation platforms (top right), the framework produces a Pareto front of resources (bottom right) along with the corresponding implementations (bottom).}
    \label{fig:informal}
    \vspace{-5mm}
\end{figure}

\paragraph{Limitations of existing approaches}
Existing work addresses important parts of this problem, but not the full loop. 
Traditional EDA flows, including high-level synthesis, accelerator compilers, and platform compilation, automate portions of hardware implementation, yet they do not provide a closed co-design loop between the neural network and its hardware realization~\cite{regleg,FlexLearn,zcomp}. 
Hardware-aware neural architecture search and accelerator design space exploration tools partly closes this gap by optimizing network architectures under hardware-related constraints~\cite{howard2019searching,wu2019fbnet,wan2020fbnetv2,timeloop,accelergy,zigzag,stream}, but in most cases the hardware platform is fixed or represented by a fixed proxy, and the
fabrication layer is ignored entirely. 
More recent co-exploration approaches, such as NAAS~\cite{naas} and DANCE~\cite{dance}, jointly search networks, accelerators, and mappings. 
Their limitation is different: the co-design methodology is tightly coupled to a particular joint search algorithm and design space, so improving one solver or replacing one model typically requires reworking the full exploration pipeline.

\paragraph{Three open challenges}
Three obstacles stand in the way of a more general solution.
The first is \emph{evaluation cost}: repeated exploration across network,
mapping, and manufacturing choices is infeasible if each candidate requires
detailed hardware design and verification, which is why prior work relies on
statistical predictors, black-box models, and coarse-grained
estimators~\cite{AdaBoost,gem5,MAESTRO,9218632}.
These surrogates are necessary but are typically introduced as standalone tools
rather than as elements of a formally composable co-design framework.
The second is \emph{uncertainty}: neural-network training is stochastic,
heuristic mapping algorithms are stochastic, and fabrication outcomes depend on process variation; yet these effects are usually collapsed into point estimates
or treated as nuisance variables, rather than as resources that can be explicitly
managed.
The third is \emph{modularity}: because existing methodologies couple the global
objective tightly to specific solvers and search spaces, improving any single
component, i.e., a training algorithm, a mapping heuristic, or a yield model, requires restructuring the entire exploration pipeline, making incremental
refinement unnecessarily expensive.
Recent multi-objective exploration frameworks and industrial co-design efforts
further highlight the need for a methodology that addresses all three obstacles
simultaneously~\cite{boom,du2025chiplever}.

\paragraph{This paper with co-design theory}
Monotone co-design theory offers a formal and compositional framework that addresses those challenges ~\cite{censi2015mathematical,zardiniCoDesignComplexSystems2023,censi2022}.
It has enabled holistic design of hardware/software architectures across robotics and control~\cite{zardiniecc21,zardiniTaskdrivenModularCodesign2022,milojevic2025codei}, heterogeneous robot fleets \cite{stralzTaskDrivenCoDesignHeterogeneous2026}, transportation~\cite{zardini2022co}, and automotive~\cite{neumann2024co}.
Together with the distributional extension~\cite{huangDistributionalUncertaintyAdaptive2026}, we formulate \glsnnp
development as the composition of five interoperable design blocks: network
selection and training, chip design through hardware mapping, fabrication
prediction, migration, and computation distribution planning.

Each block is modeled as a \glsmdpi.
The key idea is an interface-based decomposition: an \glsmdpi declares only what
\functionality{functionalities} a block can deliver and what \resource{resources}
it consumes, while the algorithm that populates its implementation set remains
internal and invisible to the rest of the system.
The co-design framework therefore interacts with each block exclusively through
this \functionality{functionality}--\resource{resource} interface, asking
\emph{what} a block can achieve for a given budget, not \emph{how} it achieves
it.
The practical payoff of this separation is that any block can be improved
without structural changes to the surrounding co-design diagram: swapping a
training algorithm, refining a hardware surrogate, or substituting a yield model
affects only the block's internal implementation set; all other blocks continue
to consume the same interface unchanged.
This directly resolves the modularity challenge and mitigates the evaluation-cost challenge: the
framework's architecture is stable across algorithmic improvements, and
offline-profiled surrogates can be slotted in without re-engineering the
composition.

\begin{figure}[t] 
    \centering
    \includegraphics[width=\columnwidth]{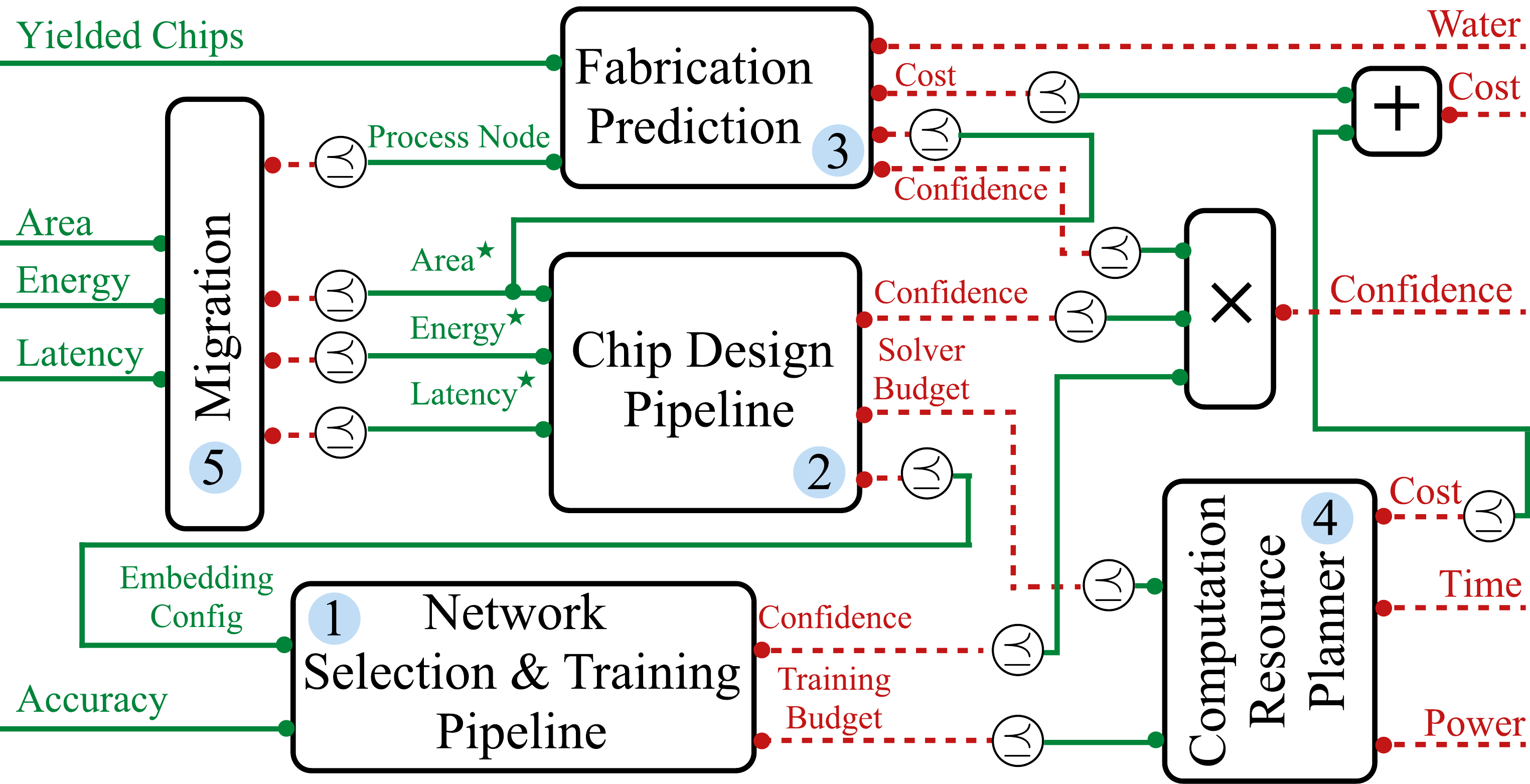} 
    \caption{The software-to-fabrication co-design diagram of neural network processors.}
    \vspace{-3mm}
    \label{fig:whole}
\end{figure}

To capture the stochasticity present in training, mapping, and fabrication, we
extend the \glsmdpi formalism to distributional uncertainty and introduce
\resource{Confidence}, defined as the inverse of success probability, as an
explicit, optimizable \resource{resource}.
This reframing converts a binary feasibility question
(\emph{``can these requirements be met?''}) into a continuous design trade-off
(\emph{``at what confidence level, and at what additional cost?''}), placing
reliability on equal footing with \resource{Time}, \resource{Power}, 
\resource{Cost}, and \resource{Water} consumption in the Pareto optimization.
Under this formulation, the \glsnnp co-design problem becomes: given
application-level requirements on \functionality{Accuracy},
\functionality{chip area}, \functionality{energy}, \functionality{latency}, and
\functionality{Yielded Chips}, select a neural network architecture, a hardware
mapping strategy, a fabrication configuration with its process node, and the supporting compute platform
so as to minimize \resource{resource} expenditure while satisfying the required
\functionality{functionalities} at the specified confidence level.
The resulting framework returns Pareto-optimal implementations together with their
associated confidence levels.
\Cref{fig:whole} summarizes the resulting software-to-fabrication co-design loop.

\paragraph{Statement of contribution}
The value of the framework is demonstrated through three case studies, each
targeting one of the challenges above.
First, we solve end-to-end \glsnnp co-design problems under multiple
application scenarios and obtain Pareto fronts that expose the trade-offs among
\resource{Cost}, \resource{Time}, \resource{Power}, \resource{Water} consumption, and \resource{Confidence},
demonstrating co-design across the full hierarchy from training to wafer
fabrication.
Second, we validate the distributional models for all three stochastic blocks and
show that \resource{Confidence} functions as a continuously tunable design knob
rather than a post hoc diagnostic, directly addressing the uncertainty challenge.
Third, we demonstrate algorithm--framework decoupling by enriching the
implementation set of the chip-design block with the implementations of a second mapping solver without modifying the surrounding
co-design diagram, confirming the modularity guarantee.

\paragraph{Organization of the manuscript}
\cref{sec:background} reviews monotone co-design theory, the \glsmdpi formalism and develops the probabilistic extension for stochastic and uncertain design blocks.
\cref{sec:co-design-model-nn-chips} instantiates the five \glspl{abk:mdpi} for the \glsnnp co-design
problem.
\cref{sec:simulation results} presents the three case studies, and \cref{sec:conclusion} concludes the paper.
The appendices collect the general stochastic co-design models, formal statement and proof of the surrogate lower bound, and the details of the chip design surrogate.

\section{Theoretical Background and Preliminaries}
\label{sec:background}
We state some notations and preliminaries on orders, then recall the fragment of monotone co-design theory used throughout the paper with uncertainty. 
The objective is to formalize, in an implementation-independent way, the trade-off between required \functionality{functionalities} and consumed \resource{resources}. 
We first define design problems, then introduce the composition operators that assemble larger systems from smaller ones, and finally make implementation choices explicit through \glspl{abk:mdpi} and introduce models with uncertainty.
A broader treatment of monotone co-design theory can be found in~\cite{zardiniCoDesignComplexSystems2023} and distributional uncertainty is introduced in~\cite{huangDistributionalUncertaintyAdaptive2026}.
Detailed proofs and additional materials that are not in the main text or references are in the appendices.

\subsection{Notations and Preliminaries on Orders}\label{subsec:math-preliminary}

We write~$f \colon A \to B$ for a function from~$A$ to~$B$, with its action on elements written~$m_A \mapsto f(m_A)$.
The \emph{Cartesian product}~$A \times B$ of sets has the tuples~$\maketup{m_A, m_B}$, with~$m_A \in A$ and~$m_B \in B$, as elements.
\emph{A partial order} compares elements of a set, and some pairs of elements may be incomparable.

\begin{definition}[Poset]
A \emph{\glsdefhere{\gls}{poset}} is a tuple~$\mathcal{P} = \maketup{P, \preceq_\mathcal{P}}$, where~$P$ is a set and~$\preceq_\mathcal{P}$ is a partial order, a relation that is reflexive, transitive, and antisymmetric.
If clear from context, we use~$P$ for a \gls{abk:poset} and~$\preceq$ for its order.
\end{definition}

\Glsplposet can model metrics other than numbers.
For example, the sets of network layer types that chips support form a \glsposet ordered by inclusion.
A chip that supports~$\{\text{conv}, \text{attention}\}$ can also behave like one that supports~$\{\text{conv}\}$, so~$\{\text{conv}\} \preceq \{\text{conv}, \text{attention}\}$.
A chip that supports~$\{\text{conv}\}$ \emph{cannot} behave like one that supports~$\{\text{attention}\}$, so~$\{\text{attention}\} \not \posetleq \{\text{conv}\}$.

The \emph{product} of \glspl{abk:poset}~$\mathcal{P}$ and~$\mathcal{Q}$ is the \glsposet~$\maketup{P \times Q, \preceq_{\mathcal{P} \times \mathcal{Q}}}$ that compares tuples componentwise:
\begin{equation*}
    \maketup{\poselxP, \poselxQ} \preceq_{\mathcal{P} \times \mathcal{Q}} \maketup{\poselyP, \poselyQ} \Leftrightarrow (\poselxP \preceq_{\mathcal{P}} \poselyP) \wedge (\poselxQ \preceq_\mathcal{Q} \poselyQ).
\end{equation*}

The \emph{opposite} of~$\mathcal{P}$ is the \glsposet~$\mathcal{P}^\op \defeq \maketup{P, \preceq_{\mathcal{P}}^\op}$ with the reversed order: $\poselxP \preceq_{\mathcal{P}}^\op \poselyP \Leftrightarrow \poselyP \preceq_{\mathcal{P}} \poselxP$.

A single metric, such as chip area or energy per inference, may be totally ordered, but several metrics together form a partial order through the product.
For example, a chip with area~$a$ in~\unit{\square\milli\metre} and energy~$e$ per inference in~\unit{\milli\joule} requires the resource tuple~$\maketup{a, e}$ in~$\mathbb{R}_+ \times \mathbb{R}_+$, each factor with the usual order~$\leq$.
Under the budget~$\maketup{3, 5}$, a chip requiring~$\maketup{2, 5}$ can be used, but one requiring~$\maketup{2, 6}$ cannot.

The \emph{upper closure} of a subset~$\subsetXP \subseteq \posetP$ contains all elements of~$\posetP$ that are greater than or equal to some element of~$\subsetXP$:
\begin{equation*}
    \upperClosure{\subsetXP} \defeq \setWithArg{\poselxP \in \posetP}{\exists \poselyP \in \subsetXP : \poselyP \posetleq_\posetP \poselxP}.
\end{equation*}
A subset~$\subsetXP \subseteq \posetP$ is an \emph{upper set} if~$\upperClosure{\subsetXP} = \subsetXP$.
Upper sets capture which budgets suffice.
If we can choose between chips requiring~$\maketup{2, 5}$ and~$\maketup{3, 4}$, the sufficient budgets form the upper set~$\upperClosure{\{\maketup{2, 5}, \maketup{3, 4}\}}$.
For every budget in this set, such as~$\maketup{2.5, 6}$, there exists a usable chip, here the one requiring~$\maketup{2, 5}$.

\begin{definition}[Monotone map]
A map~$f \colon \posetP \to \posetQ$ between \glspl{abk:poset}~$\mathcal{P}$ and~$\mathcal{Q}$ is \emph{monotone} if~$\poselxP \preceq_\mathcal{P} \poselyP \Rightarrow f(\poselxP) \preceq_\mathcal{Q} f(\poselyP)$.
Monotonicity is preserved by composition and products.
\end{definition}

Monotone maps capture the performance of many processes.
For instance, given a more complicated neural network (with more layers or parameters), the minimum energy required to run it is larger than or equal to the minimum energy required to run a simpler network.

However, monotone maps are not enough to capture trade-offs in general \glsplposet.
If we consider both area and energy, there might be various chips that can run the same neural network with non-comparable area and energy budgets.
Those trade-offs are captured in \emph{monotone co-design}.

\vspace{-4mm}
\subsection{Monotone Co-Design Theory}

\begin{definition}[\glsdefhere{\Glspl}{dp}]
Given \glsplposet of functionalities~$\functionality{F}$ and resources~$\resource{R}$, a \glsdp is an upper set of the product \glsposet~$\functionality{F}^{\mathrm{op}} \times \resource{R}$. 
We denote the set of such \glspldp by~$\mathcal{DP}\{\functionality{F}, \resource{R}\}$.
Given a \glsdp $\dprb$, a pair $\maketup{x_{\functionality{F}}, x_{\resource{R}}}$ of functionality $x_{\functionality{F}}$ and resource $x_{\resource{R}}$ is \emph{feasible} if $\maketup{x_{\functionality{F}}, x_{\resource{R}}} \in \dprb$.
\end{definition}

\Glspldp contain the feasible pairs of \functionality{functionalities} and \resource{resources}.
The upper set condition captures the intuition that if a resource~${x_{\resource{R}}}$ suffices for functionality~${x_{\functionality{F}}}$, it also suffices for any lesser functionality~${x_{\functionality{F}}}' \preceq {x_{\functionality{F}}}$, and any greater resource~${x_{\resource{R}}}' \succeq {x_{\resource{R}}}$ also suffices for~${x_{\functionality{F}}}$.

For example, in the co-design of a network and its chip, let the \functionality{functionality} be the inference accuracy in~$[0, 1]$ and the \resource{resources} the area and energy of the chip in~$\mathbb{R}_+ \times \mathbb{R}_+$.
If one design reaches accuracy~$0.90$ with~$\maketup{3, 4}$ and another reaches~$0.85$ with~$\maketup{2, 5}$, the \glsdp of this choice is~$\upperClosure{\{\maketup{0.90, \maketup{3, 4}}, \maketup{0.85, \maketup{2, 5}}\}}$ in~$\functionality{F}^{\mathrm{op}} \times \resource{R}$.
The pair~$\maketup{0.88, \maketup{3.5, 4}}$ is feasible through the first design, whereas~$\maketup{0.90, \maketup{2, 5}}$ is infeasible, since only the first design reaches~$0.90$ and it needs more area.

Complex systems are composed of simpler sub-systems, and we formalize these compositions as operations on \glspldp.

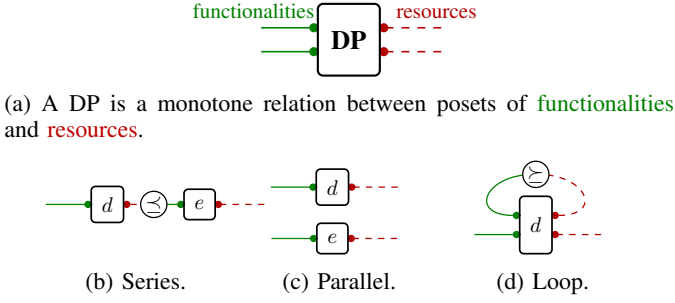
\begin{figure}[tb]

\begin{subfigure}{\columnwidth}
    \begin{center}
    \begin{tikzpicture}[DP]
            \node[dp={2}{2}] (cnt) {DP};
            \draw[runconn, runame={resources}, relres=above,posres=0.9] (cnt_res1){};
            \draw[runconn, runame={}, relres=above,posres=0.9] (cnt_res2){};
            \draw[funconn, funame={functionalities},relfun=above,posfun=1.15] (cnt_fun1){};
            \draw[funconn, funame={},relfun=above,posfun=1.15] (cnt_fun2){};
\end{tikzpicture}
    \subcaption{A \gls{abk:dp} is a monotone relation between \glsplposet of \functionality{functionalities} and \resource{resources}. \label{fig:mathcodesign}}
    \end{center}
\end{subfigure}

\begin{center}
\begin{subfigure}[b]{0.38\columnwidth}
  \centering
  \scalebox{0.8}{\begin{tikzpicture}[DP]
    \node[dp={1}{1}] (f) {$d$};
    \node[dp={1}{1}, right=1cm of f] (g) {$e$};
    \draw[rconn, rcname={}, fcname={}] (f_res1)  to (g_fun1);
    \draw[runconn, runame={}] (g_res1);
    \draw[funconn, funame={}] (f_fun1);
\end{tikzpicture}}
  \subcaption{Series.}
\end{subfigure}
\begin{subfigure}[b]{0.25\columnwidth}
  \centering
  \scalebox{0.8}{\begin{tikzpicture}[DP]
    \node[dp={1}{1}] (f) {$d$};
    \node[dp={1}{1}, below=0.3cm of f] (g) {$e$};
    \draw[runconn, runame={}] (f_res1){};
    \draw[runconn, runame={}] (g_res1){};
    \draw[funconn, funame={}] (f_fun1){};
    \draw[funconn, funame={}] (g_fun1){};
\end{tikzpicture}}
  \subcaption{Parallel.}
\end{subfigure}
\begin{subfigure}[b]{0.25\columnwidth}
  \centering
  \scalebox{0.8}{\begin{tikzpicture}[DP]
    \node[dp={2}{2}] (f) {$d$};
    \draw[runconn, runame={}] (f_res2){};
    \draw[funconn, funame={}] (f_fun2){};
    \draw[rconn,rcname={},fcname={},feedback=1,loos=3] (f_res1) -- ($(f)+(0,4)$) |- (f_fun1);
\end{tikzpicture}}
  \subcaption{Loop.}
\end{subfigure}

\label{fig:diagrams}
\caption{\Glspldp can be composed in different ways.}
\label{fig:dp_def}
\end{center}
\vspace{-7mm}
\end{figure}

\begin{definition}[Composition operations for \glspl{abk:dp}]\label{def:dp-compositions}
    The following operations construct new \glspl{abk:dp} from existing ones.
    
    \noindent \emph{Series}: Given \glspl{abk:dp} $\dprba \in \dpOf{\posetP}{\posetQ}$ and $\dprbb \in \dpOf{\posetQ}{\posetR}$, their series connection $\dprba \mthenMathbin \dprbb \in \dpOf{\posetP}{\posetR}$ is defined as
        \[            
        \{\maketup{\poselxP, \poselxRnoRes} \mid
            \exists \poselxQ : \maketup{\poselxP, \poselxQ} \in \dprba
            \text{ and }\maketup{\poselxQ, \poselxRnoRes} \in \dprbb\}.
        \]
    This models situations where $\dprba$ uses the functionalities provided by $\dprbb$ as its resources. \\
    \emph{Parallel}: For $\dprba \in \dpOf{\posetP}{\posetQ}$ and $\dprba' \in \dpOf{\posetP'}{\posetQ'}$, their parallel connection $\dprba \stackMathbin \dprba' \in \dpOf{\posetP \times \posetP'}{\posetQ \times \posetQ'}$ is
        \[
            \{\maketup{\maketup{\poselxP,\poselxP'}, \maketup{\poselxQ,\poselxQ'}} \mid 
            \maketup{\poselxP, \poselxQ} \in \dprba
            , \maketup{\poselxP', \poselxQ'} \in \dprba' \}.
        \]
    It represents two non-interacting systems. \\
    \emph{Feedback/Trace}: For $\dprb \in \dpOf{\posetP \times \posetR}{\posetQ \times \posetR}$, its trace $\traceOf{\dprb} \in \dpOf{\posetP}{\posetQ}$ is defined as
        \[
            \{ \maketup{\poselxP, \poselxQ} \mid
            \exists \poselxRnoRes : \maketup{\maketup{\poselxP,\poselxRnoRes}, \maketup{\poselxQ,\poselxRnoRes}} \in \dprb \}.
        \]
    This models the case where functionalities provided by $\dprb$ are used as its own resources. \\
    \emph{Union} and \emph{intersection}: Given $\dprba,\dprbb \in \dpOf{\posetP}{\posetQ}$, their union $\dprba \union \dprbb \in \dpOf{\posetP}{\posetQ}$ is defined by
        \[
            \{\maketup{\poselxP,\poselxQ} \mid \\
            \maketup{\poselxP, \poselxQ} \in \dprba
            \text{ or }\maketup{\poselxP, \poselxQ} \in \dprbb \}.
        \]
    Union expresses a free choice between~$\dprba$ and~$\dprbb$.
    The intersection~$\dprba \intersection \dprbb \in \dpOf{\posetP}{\posetQ}$, defined as
        \[
            \{\maketup{\poselxP,\poselxQ} \mid
            \maketup{\poselxP, \poselxQ} \in \dprba
            \text{ and }\maketup{\poselxP, \poselxQ} \in \dprbb \},
        \]
    requires satisfying both.
    Union and intersection extend to sets of \glspl{abk:dp}, for instance~$\union\setWithIndex{\dprb_i}{i \in I}$.
\end{definition}

\begin{definition}[\glsdefhere{\Gls}{mdpi}]\label{def:mdpi}
    Designers care about which \functionality{functionality}/\resource{resource} pairs are feasible and also about which design choices realize them.
    To reason about such choices, co-design introduces \emph{\implementation{implementations}}, colored in \implementation{orange}.
    Given \glsplposet~$\functionality{F}$ and~$\resource{R}$, an \glsmdpi is a tuple~$\maketup{\implementation{I}, \text{realize}}$ of a set of \implementation{implementations}~$\implementation{I}$ and a map~$\text{realize} \colon \implementation{I} \to \mathcal{DP}\{\functionality{F},\resource{R}\}$, where~$\dprb_{\implementation{i}} \defeq \text{realize}(\implementation{i})$ is the \functionality{functionality}/\resource{resource} trade-off provided by~$\implementation{i}$.
    It induces a \glsdp as the free choice among all implementations, $\dprb = \union \{\dprb_{\implementation{i}}\}_{\implementation{i} \in \implementation{I}}$, so a pair~$\maketup{x_{\functionality{F}},x_{\resource{R}}}$ is feasible in~$\dprb$ if and only if some implementation realizes it.
\end{definition}

The example above is an \glsmdpi with~$\implementation{I} = \{\implementation{i}_1, \implementation{i}_2\}$, $\text{realize}(\implementation{i}_1) = \upperClosure{\{\maketup{0.90, \maketup{3, 4}}\}}$, and~$\text{realize}(\implementation{i}_2) = \upperClosure{\{\maketup{0.85, \maketup{2, 5}}\}}$, where each implementation gives a single pair.
An implementation can also be a design process, such as a hardware-mapping algorithm, and then give a trade-off with multiple Pareto solutions, for instance different area and energy values for different accuracy objectives.

\begin{definition}[Queries for \glsplmdpi]\label{def:mdpi-queries}
    For an \glsmdpi~$\langle\implementation{I}, \text{realize}\rangle$ with functionality and resource \glsplposet~$\functionality{F}$ and~$\resource{R}$, the \emph{Fix \functionality{functionalities} minimize \resource{resources}} query takes a set of required functionalities~$A_{\functionality{F}} \subseteq \functionality{F}$ and returns all pairs of resources~$x_{\resource{R}} \in \resource{R}$ and design choices~$\implementation{i} \in \implementation{I}$ such that~$\maketup{x_{\functionality{F}},x_{\resource{R}}}$ is feasible in~$\text{realize}(\implementation{i})$ for every~$x_{\functionality{F}} \in A_{\functionality{F}}$.
\end{definition}

Solving such queries for composite systems is in general a non-convex,
non-continuous, and potentially combinatorial multi-objective optimization problem.
A fundamental property of the formalism is that solutions compose in exactly
the same way as problems: the query result for a composed system is obtained
by composing the query results of its constituent blocks, following the same
series, parallel, and feedback operations that assemble the co-design diagram.
For systems containing feedback, this takes the form of a fixed-point
iteration: under standard assumptions of complete \glsplposet and Scott-continuity,
Kleene's fixed-point theorem guarantees convergence to the Pareto-optimal
solution set or certifies infeasibility~\cite{censi2015mathematical,%
zardiniCoDesignComplexSystems2023}.
The computational cost scales linearly in the number of implementations in
each block (up to antichain operations), rather than combinatorially across
blocks, avoiding the exponential blowup that would follow from exhaustive
enumeration of all cross-block design combinations~\cite{censi2015mathematical,%
zardiniCoDesignComplexSystems2023}.

\subsection{Co-Design with Uncertainty by Confidence}
\label{sec:probabilistic-framework}
Many co-design blocks in the present workflow are stochastic: neural-network training depends on random initialization and mini-batch order, heuristic hardware mapping depends on randomized exploration, and fabrication depends on process variation.
Building on the distributional co-design framework of~\cite{huangDistributionalUncertaintyAdaptive2026}, we model stochastic components by adding \resource{Confidence}, the inverse of probability, as a resource.

The main query considered in this paper fixes required \functionality{functionalities} and minimizes \resource{resources}. 
A success probability~$\pi \in [0,1]$ is monotone in the opposite direction, so we encode it as the \resource{Confidence resource}~$\pi^{-1} \in [1,+\infty]$. 
Smaller values correspond to higher reliability. 

\subsubsection{\texorpdfstring{Budgeted Stochastic \glspl{abk:mdpi}}{Budgeted Stochastic MDPIs}}
\label{sec:stochastic-design-blocks}
In general, a distributional uncertain \gls{abk:mdpi} assigns a random \gls{abk:dp} to each implementation~\cite{huangDistributionalUncertaintyAdaptive2026}.
The stochastic blocks of this paper have a simpler structure.
An iterative algorithm, such as \gls{abk:sgd} training or a randomized mapping search, receives a computational budget and returns a result whose quality depends on random outcomes, such as the initialization and the mini-batch order.
Over all random outcomes, the performance of such an algorithm is a deterministic map.

\begin{definition}[Budgeted stochastic \gls{abk:mdpi}]
\label{def:budgeted-stochastic}
Let~$\implementation{I}$ be an implementation set, $\functionality{F}$ and~$\resource{R}$ \glsplposet, and~$\maketup{\Omega,\outcomeDist}$ a probability space with a distribution.
A \emph{budgeted stochastic \gls{abk:mdpi}} is a measurable map~$\Phi : \implementation{I} \times \resource{R} \times \Omega \to \functionality{F}$, where~$\Phi(\implementation{i},x_{\resource{R}},\omega)$ is the functionality delivered by implementation~$\implementation{i}$ under resource budget~$x_{\resource{R}}$ and random outcome~$\omega$.
\end{definition}

\subsubsection{\texorpdfstring{Surrogate Model for Budgeted Stochastic \glspl{abk:mdpi}}{Surrogate Model for Budgeted Stochastic MDPIs}}
\label{sec:surrogate-model}
The sample space~$\Omega$ is often intractable.
In \gls{abk:sgd}, for instance, the sequence of mini-batches is generated online and is not modeled explicitly.
We therefore work with the induced distribution over \functionality{functionalities}.
Through the map~$\omega \mapsto \Phi(\implementation{i},x_{\resource{R}},\omega)$ and the distribution~$\outcomeDist$, each implementation~$\implementation{i}$ and budget~$x_{\resource{R}}$ induce a conditional distribution~$\inducedDist{\cdot}{\implementation{i},x_{\resource{R}}}$ over~$\functionality{F}$, which Appendix~\ref{app:stochastic-mdpi} defines formally.
The surrogate models of \cref{sec:co-design-model-nn-chips} are parametric forms of these distributions fit from data.
The probability of meeting a functionality requirement~$x_{\functionality{F}}$ under implementation~$\implementation{i}$ and resource budget~$x_{\resource{R}}$ is then
\begin{equation}\label{eq:success-probability-general}
    p_\Phi(\implementation{i},x_{\resource{R}};x_{\functionality{F}})
    =
    \inducedDist{\upperClosure{\makeset{x_{\functionality{F}}}}}{\implementation{i},x_{\resource{R}}},
\end{equation}
and the corresponding confidence-augmented \gls{abk:mdpi} is
\begin{equation}\label{eq:confidence-augmented-mdpi}
    \implementation{i}
    \mapsto
    \makeset{
        \maketup{x_{\functionality{F}},\maketup{x_{\resource{R}},\pi^{-1}}}
        \mid
        \pi \le \sup_{x_{\resource{R}}' \preceq x_{\resource{R}}} p_\Phi(\implementation{i},x_{\resource{R}}';x_{\functionality{F}})
    }.
\end{equation}

\begin{remark}[Best-so-far policy]
\label{rmk:best_so_far}
Many iterative algorithms are not monotone in their raw performance for a fixed random outcome.
Network training, for example, may overfit and lose test accuracy in the long run.
The supremum over~$x_{\resource{R}}' \preceq x_{\resource{R}}$ in \cref{eq:confidence-augmented-mdpi} implicitly assumes a best-so-far policy that keeps the best result obtained so far.
If training reaches \qty{80}{\percent} test accuracy after 10 epochs but overfits to \qty{70}{\percent} after 20 epochs, the policy returns \qty{80}{\percent} for any budget~$x_{\resource{R}} \geq 10$ epochs, since the model at 10 epochs can always be deployed.
If~$\functionality{F}$ is totally ordered, the policy can also be built into~$\Phi$ itself.
This can induce more structured surrogate models as in the neural network training (\cref{sec:nn-architectures}).
\end{remark}

\begin{remark}[Priority as one implementation and monotonicity]
\label{rmk:priority-as-imp}
When~$\functionality{F}$ is only partially ordered, an algorithm may need a priority to trade off \functionality{functionalities}.
For the hardware-mapping algorithm of \cref{sec:hw-mapping} with \functionality{Latency} prioritized, more computation may give better \functionality{Latency} but worse \functionality{Energy}.
The map~$\Phi$ is then not monotone in~$\resource{R}$, but the best-so-far policy still makes \cref{eq:confidence-augmented-mdpi} a valid \glsmdpi.
\end{remark}

In this surrogate, the budget~$x_{\resource{R}}$ is chosen \emph{before} the random outcome~$\omega$ is known.
In practice, the designer may choose how much resource, for instance computation, to spend \emph{after} partly observing the random outcome and thus the realized functionality.
Hence, the surrogate model provides a \emph{lower bound} on the probability of satisfying the functionality requirement~$x_{\functionality{F}}$ under implementation~$\implementation{i}$ and resource budget~$x_{\resource{R}}$.
We provide a formal statement and proof in Appendix~\ref{app:surrogate-proof}.

\section{\texorpdfstring{\Gls{abk:mdpi}}{MDPI} Model of Neural Network-Processor Co-Design}
\label{sec:co-design-model-nn-chips}

Using the probabilistic framework of \cref{sec:probabilistic-framework}, we instantiate the end-to-end neural-network-processor co-design problem as the composition of five \glspl{abk:mdpi}: \bluecircled{1}\textbf{network selection and training}, \bluecircled{2}\textbf{chip design} through hardware mapping, \bluecircled{3}\textbf{fabrication prediction},  \bluecircled{4}\textbf{computation planning}, and technology \bluecircled{5}\textbf{migration}. The resulting diagram is shown in \cref{fig:whole}. The system-level query fixes application requirements on \functionality{Accuracy}, \functionality{Chip Area}, \functionality{Chip Energy}, \functionality{Chip Latency}, and \functionality{Yielded Chips}; the solver returns Pareto-minimal \resource{Water}, \resource{Cost}, \resource{Time}, \resource{Power}, and \resource{Confidence}. 
Throughout this section, lower physical values of area, energy, and latency are preferred. We therefore equip these physical metrics with the reversed order, or equivalently treat a smaller physical value as a larger \functionality{functionality} in the co-design sense.

The five blocks communicate only by functionality-resource interfaces. 
The \bluecircled{1}\textbf{Network Selection and Training Pipeline} provides a deployment embedding configuration and an accuracy guarantee. 
The \bluecircled{2}\textbf{Chip Design Pipeline} consumes the embedding configuration and provides chip-level area, energy, and latency. 
The \bluecircled{3}\textbf{Fabrication Prediction} block provides the number of yielded chips together with the process node, consuming the chip area at the reference node, water consumption, and monetary cost. 
The \bluecircled{4}\textbf{Computation Distribution Planner} provides the computational budgets required by the training and mapping blocks, while consuming wall-clock time, power, and monetary cost.
The \bluecircled{5}\textbf{Migration} block takes the process node together with the chip area, energy, and latency at the reference node, and returns these three metrics at the given process node.

Probabilistic blocks expose the resource \resource{Confidence} and the system-level confidence is the product of block confidences.

\begin{remark}[Independence of success probabilities]
    \label{rmk:independence-success-probabilities}
    In a query, the solver selects one implementation per block and one value for each \functionality{functionality} and \resource{resource}.
    The success probability of a stochastic block is the probability that its implementation, under the selected \resource{resource} budget, delivers the selected \functionality{functionality} requirement.
    This event depends only on the block's own random outcome, and these outcomes are independent across blocks, so the events that the blocks meet their requirements are independent.
    The success probability of the solution is therefore the product of the block success probabilities, and the system-level \resource{Confidence} is the product of block \resource{Confidences}.
    Dependences between blocks are captured by the \functionality{functionality}--\resource{resource} interfaces, such as Chip Area between chip design and fabrication.
    For example, the solver may select \qty{4}{\square\milli\metre} on this interconnection, choose a mapping strategy giving an area at most \qty{4}{\square\milli\metre} with probability~$0.9$, and select a fabrication strategy that yields at least 100 working chips of a \qty{4}{\square\milli\metre} die with probability~$0.8$.
    Both probabilities refer to the selected area, not to the area of the mapping found in a particular run: a larger area is already counted as a failure in the $0.9$, and a smaller area yields at least as many working chips, so the product never overstates the success probability.
\end{remark}
This interface-level description enables modularity. 
A block may be refined internally, for example by replacing a training surrogate, a mapping solver, or a fabrication yield model, without changing the system-level co-design diagram as long as the same \functionality{functionality}-\resource{resource} interface is preserved.

\subsection{Network Selection and Training Pipeline}
\label{sec:nn-architectures}
Networks induce trade-offs.
A larger network usually reaches a higher accuracy, but it needs more arithmetic operations and memory for training and inference.
Evaluating this trade-off is expensive: one must train the network, measure its accuracy, and consider its deployment on hardware.

To address this problem, the \bluecircled{1}\textbf{network selection and training} block captures the trade-off between deployment-relevant network properties, training budget, and attainable accuracy with a surrogate model fitted from datasets. 
Unlike a conventional neural architecture search objective that optimizes only accuracy or accuracy under a fixed proxy constraint, this block returns both the embedding configuration needed by the hardware block and a probability of achieving a requested accuracy under a finite training budget.

\subsubsection{Abstract Model}

\begin{definition}[Layers and candidate networks]
\label{def:layers-and-networks}
Let \(\mathcal{O}\) be a finite set of layer operator types, such as convolution, pooling, and normalization.
A layer is a tuple \(\mathtt{layer}=\maketup{op,shape}\), where \(op\in\mathcal{O}\) and \(shape\) encodes tensor dimensions.
A candidate network is an ordered sequence \(\implementation{\theta}=\maketup{\mathtt{layer}_1,\ldots,\mathtt{layer}_N}\) of connected layers, and \(\implementation{\Theta}\) is the set of candidate networks.
\end{definition}

For a network \(\implementation{\theta}\), a setting \(\implementation{\trainsetting}\) in the set~\(\implementation{\trainsettingSet}\) of candidate training settings, and an epoch budget \(t\in\mathbb{N}_{+}\), training is modeled as a map
$
    \Psi_{\mathrm{train}}
    :
    \implementation{\Theta}
    \times
    \implementation{\trainsettingSet}
    \times
    \mathbb{N}_{+}
    \times
    \Omega_{\mathrm{train}}
    \to
    [0,1]
$,
whose value \(\Psi_{\mathrm{train}}(\implementation{\theta},\implementation{\trainsetting},t,\omega)\) is the best-so-far validation or test accuracy up to epoch \(t\) under random outcome \(\omega\).
The sample space \(\Omega_{\mathrm{train}}\) includes random initialization, data ordering, mini-batch sampling, and all other stochastic effects of the optimization.
Each~\(\implementation{\theta}\), \(\implementation{\trainsetting}\), and~\(t\) induce a conditional distribution~\(\inducedDist{\cdot}{\implementation{\theta},\implementation{\trainsetting},t}\) over the accuracy, as in \cref{sec:surrogate-model}, and the success probability for a requested accuracy \(\functionality{\mathfrak{a}}\) is
\begin{equation}
\label{eq:confidence-training-accuracy}
    p_{\mathrm{train}}(\implementation{\theta},\implementation{\trainsetting},t;\functionality{\mathfrak{a}})
    =
    \inducedDist{[\functionality{\mathfrak{a}},1]}{\implementation{\theta},\implementation{\trainsetting},t}.
\end{equation}

\subsubsection{Surrogate Model}

We approximate \(\inducedDist{\cdot}{\implementation{\theta},\implementation{\trainsetting},t}\) by the Gaussian surrogate
\begin{equation}
\label{eq:uncertainty-train}
    A_{\implementation{\theta},\implementation{\trainsetting},t}
    \sim
    \mathcal{N}\!\left(
        \mu_{\implementation{\theta},\implementation{\trainsetting}}(t),
        \sigma_{\mathrm{total}}^{2}
    \right),
\end{equation}
whose mean trajectory is
\begin{multline}
\label{eq:hybrid-curve}
    \mu_{\implementation{\theta},\implementation{\trainsetting}}(t)
    =
    a_{\infty,\implementation{\theta},\implementation{\trainsetting}}
    +
    A_{\implementation{\theta},\implementation{\trainsetting}}
    e^{\alpha_{\implementation{\theta},\implementation{\trainsetting}}t}
    \left(1-0.8w_{\implementation{\theta},\implementation{\trainsetting}}(t)\right)
    \\
    -
    B_{\implementation{\theta},\implementation{\trainsetting}}
    t^{\beta_{\implementation{\theta},\implementation{\trainsetting}}}
    \left(0.2+0.8w_{\implementation{\theta},\implementation{\trainsetting}}(t)\right),
\end{multline}
where
\begin{equation*}
    w_{\implementation{\theta},\implementation{\trainsetting}}(t)
    =
    \frac{1}{
        1+
        \exp\!\left(
            -s_{\implementation{\theta},\implementation{\trainsetting}}
            \frac{t-C_{\implementation{\theta},\implementation{\trainsetting}}}
                 {C_{\implementation{\theta},\implementation{\trainsetting}}}
        \right)
    }
\end{equation*}
interpolates between early exponential improvement and late power-law convergence.
The parameters are fitted with L-BFGS-B to recorded training trajectories of the candidate networks of HW-NAS-Bench~\cite{li2021hwnasbenchhardwareaware} and Accel-NASBench~\cite{accelnasbench} on CIFAR-10 with the settings of \cref{table:training settings}.
The networks of both data sets are included in the implementation set of this block.
The residual standard deviation is treated as homoscedastic and calibrated as \(\sigma_{\mathrm{total}}\approx 0.066\) for HW-NAS-Bench and \(\sigma_{\mathrm{total}}\approx 0.017\) for Accel-NASBench, so we have
\begin{equation}
\label{eq:p-train-gaussian}
    p_{\mathrm{train}}(\implementation{\theta},\implementation{\trainsetting},t;\functionality{\mathfrak{a}})
    \approx
    1-F_{\mathcal{N}}\!\left(
        \functionality{\mathfrak{a}};
        \mu_{\implementation{\theta},\implementation{\trainsetting}}(t),
        \sigma_{\mathrm{total}}^{2}
    \right).
\end{equation}

\begin{table}[tb]
\centering
\caption{Training settings used for the surrogate.}
\label{table:training settings}
\begin{tabularx}{0.7\columnwidth}{@{}lX@{}}
\toprule
\textbf{Setting} & \textbf{Value/Description} \\
\midrule
Optimizer & \glssgd \\
Momentum & 0.9 \\
Weight Decay & \(5\times 10^{-4}\) \\
Initial Learning Rate & 0.1 \\
Batch Size & 256 \\
\bottomrule
\end{tabularx}
\vspace{-4mm}
\end{table}

\subsubsection{\texorpdfstring{\gls{abk:mdpi}}{MDPI} Model}

\begin{figure}[t]
    \centering
    \includegraphics[width=0.3\textwidth]{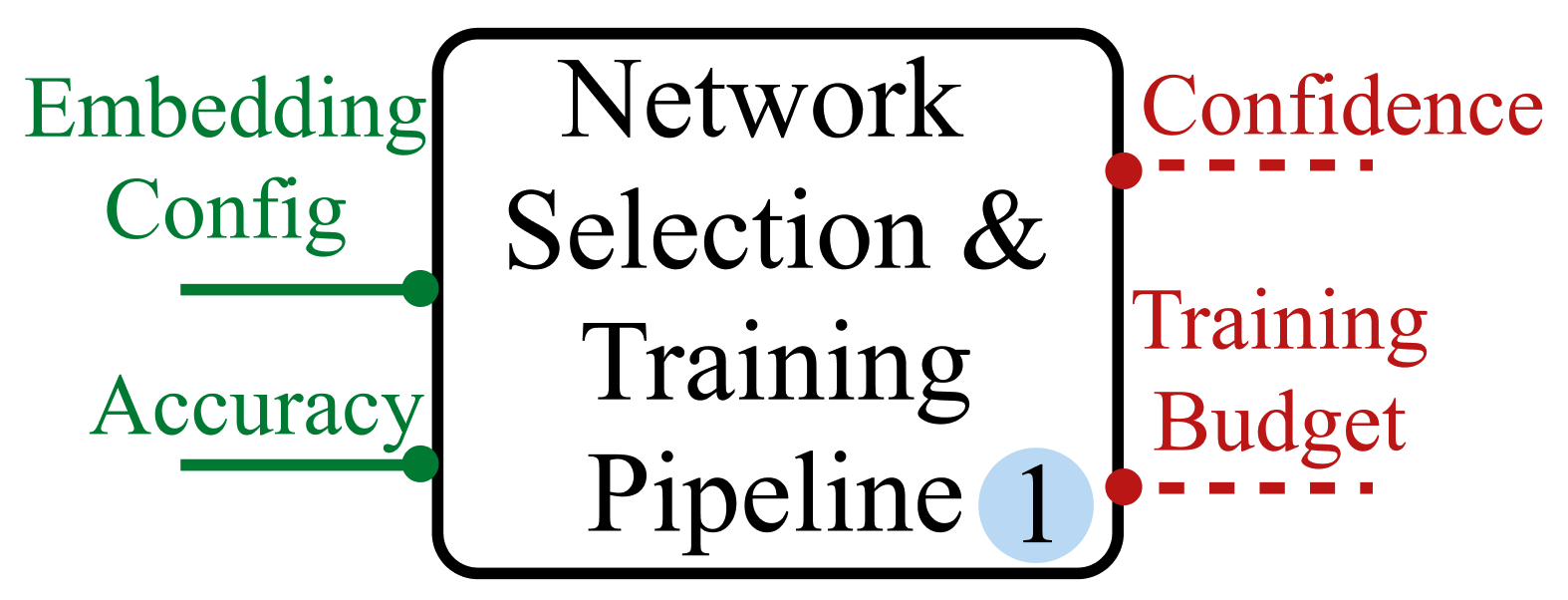}
    \caption{\texorpdfstring{\protect\bluecircled{1}}{(1)}\textbf{Network Selection and Training Pipeline} \gls{abk:mdpi}.}
    \label{fig:sw-mdpi-block}
    \vspace{-5mm}
\end{figure}

The implementation set of the training block is \(\implementation{\Theta}\times\implementation{\trainsettingSet}\times\mathbb{N}_{+}\) of networks, training settings, and epochs.
The block provides an \functionality{Embedding Configuration} \(\functionality{\ell}\in\functionality{L^\star}\) and a \functionality{Task Accuracy} \(\functionality{\mathfrak{a}}\in[0,1]\), and consumes \resource{Training Budget}~$\resource{b}_{\mathrm{train}}(\implementation{\theta},\implementation{\trainsetting},t) \in \resource{B}$ and \resource{Confidence}~$\resource{\pi^{-1}} \in [1, \infty)$.
In general, $\functionality{L^\star}$ can be a \glsposet of inference-time computing requirements, such as FLOP/s, memory, and bandwidth, with a map~$\functionality{\ell}_\mathrm{net}$ sending each network~$\implementation{\theta}$ to its requirements~$\functionality{\ell}_\mathrm{net}(\implementation{\theta}) \in \functionality{L^\star}$.
Here, we expose the network structure itself, $\functionality{\ell}_\mathrm{net}(\implementation{\theta}) = \implementation{\theta}$, with the discrete order, in which $\functionality{\ell}_1 \posetleq_{\functionality{L^\star}} \functionality{\ell}_2$ if and only if $\functionality{\ell}_1 = \functionality{\ell}_2$.
The \gls{abk:mdpi} realized by implementation \(\maketup{\implementation{\theta},\implementation{\trainsetting},t}\) is
\begin{multline}
\label{eq:network-training-pipeline-mdpi}
    \maketup{\implementation{\theta},\implementation{\trainsetting},t}
    \mapsto
    \Big\{
    \maketup{
        \maketup{\functionality{\ell},\functionality{\mathfrak{a}}},
        \maketup{\resource{b}',\resource{\pi^{-1}}}
    }
    \;\Big|\;
    \functionality{\ell}
    \posetleq_{\functionality{L^\star}}
    \functionality{\ell}_\mathrm{net}(\implementation{\theta})=\implementation{\theta},
    \\
    \resource{b}'\ge
    \resource{b}_{\mathrm{train}}(\implementation{\theta},\implementation{\trainsetting},t),
    \quad
    \resource{\pi}
    \le
    p_{\mathrm{train}}(\implementation{\theta},\implementation{\trainsetting},t;\functionality{\mathfrak{a}})
    \Big\}.
\end{multline}
The co-design framework is agnostic to the internal surrogate model.
Designers can train the surrogate with more recent datasets, use other surrogates, or include first-principles models, as long as they share the same interface.

\subsection{Chip Design Pipeline}
\label{sec:hw-mapping}

Hardware mapping also induces trade-offs among area, energy, and latency.
A mapping determines how each layer runs on the chip---dataflow, tiling, parallelization, scheduling, and data movement---and a mapping good in one metric may be bad in another.
The mapping space is discrete and high-dimensional, so gradient-based optimization is hardly possible and a search algorithm must evaluate many candidates.
Runs of a randomized search with the same budget also return mappings of different quality.

The \bluecircled{2}\textbf{chip design} block captures this trade-off.
We use a \glsga-based solver~\cite{gamma} to explore the mapping space, though any solver with the same interface can be incorporated.
Because runs with the same budget return mappings of different quality, the block returns, for a given budget, the probability of reaching the requested chip metrics rather than a single mapping.
It requires the embedding configuration~$\resource{\ell}$, the solver budget~$\resource{b}$, and \resource{Confidence}, and provides the chip metrics \functionality{\glslea}~$\functionality{\mathtt{X}}$\footnote{
In the rest of this section, \glslea denote the reciprocals of the physical values, so that a larger value is better, as in the framework of \glsmdpi.
}.
The embedding configuration is a \functionality{functionality} provided by \bluecircled{1}\textbf{Network Training Pipeline} and a \resource{resource} required here, so $\resource{\ell}$ takes the \resource{resource} color.

\subsubsection{Abstract Model}
\begin{figure}[t]
    \centering
    \includegraphics[width=0.3\textwidth]{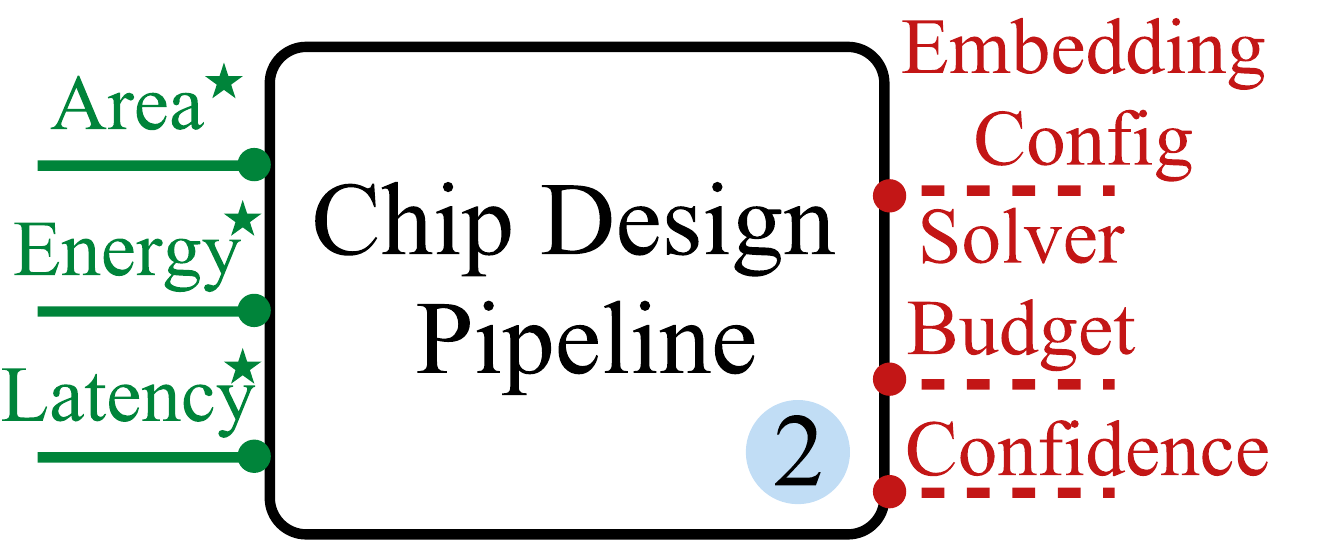}
        \caption{\bluecircled{2}\textbf{Chip Design Pipeline} as an \gls{abk:mdpi} module.}
    \label{fig:hw-mdpi-block}
\end{figure}

\begin{table}[htbp]
\centering
\caption{Solver configurations.}
\begin{tabularx}{\columnwidth}{@{}llX@{}}
\toprule
\textbf{Category} & \textbf{Parameter} & \textbf{Value/Description} \\
\midrule

\multirow{2}{*}{\textit{\glsga Solver Settings}}
& Population size & 1000 \\
& epochs          & 3000$^{\dagger}$ \\
\midrule

\multirow{2}{*}{\textit{Mapping Strategy}}
& slevel          & Spatial hierarchy level (Default: 2) \\
& parRS           & Enable parallelization \\
\midrule

\multirow{5}{*}{\textit{Hardware Constraints}}
& num\_pe         & Number of \glsxtrshortpl{abk:pe} (1024) \\
& l1\_size        & \qty{4}{\mebi\byte}$^{\dagger}$ (\glsxtrshort{abk:lone} buffer per \glsxtrshort{abk:pe}) \\
& l2\_size        & \qty{24}{\mebi\byte}$^{\dagger}$ (shared \glsxtrshort{abk:ltwo} buffer) \\
& NocBW           & \textbf{Unlimited} (Assumed sufficient) \\
& offchipBW       & \textbf{Unlimited} (Assumed sufficient) \\
\bottomrule
\end{tabularx}
\par\smallskip\raggedright\footnotesize\noindent $^{\dagger}$Upper bound of a search range.
\label{tab:hw-config}
\end{table}

\begin{figure}[t]
    \centering
    \includegraphics[width=\columnwidth]{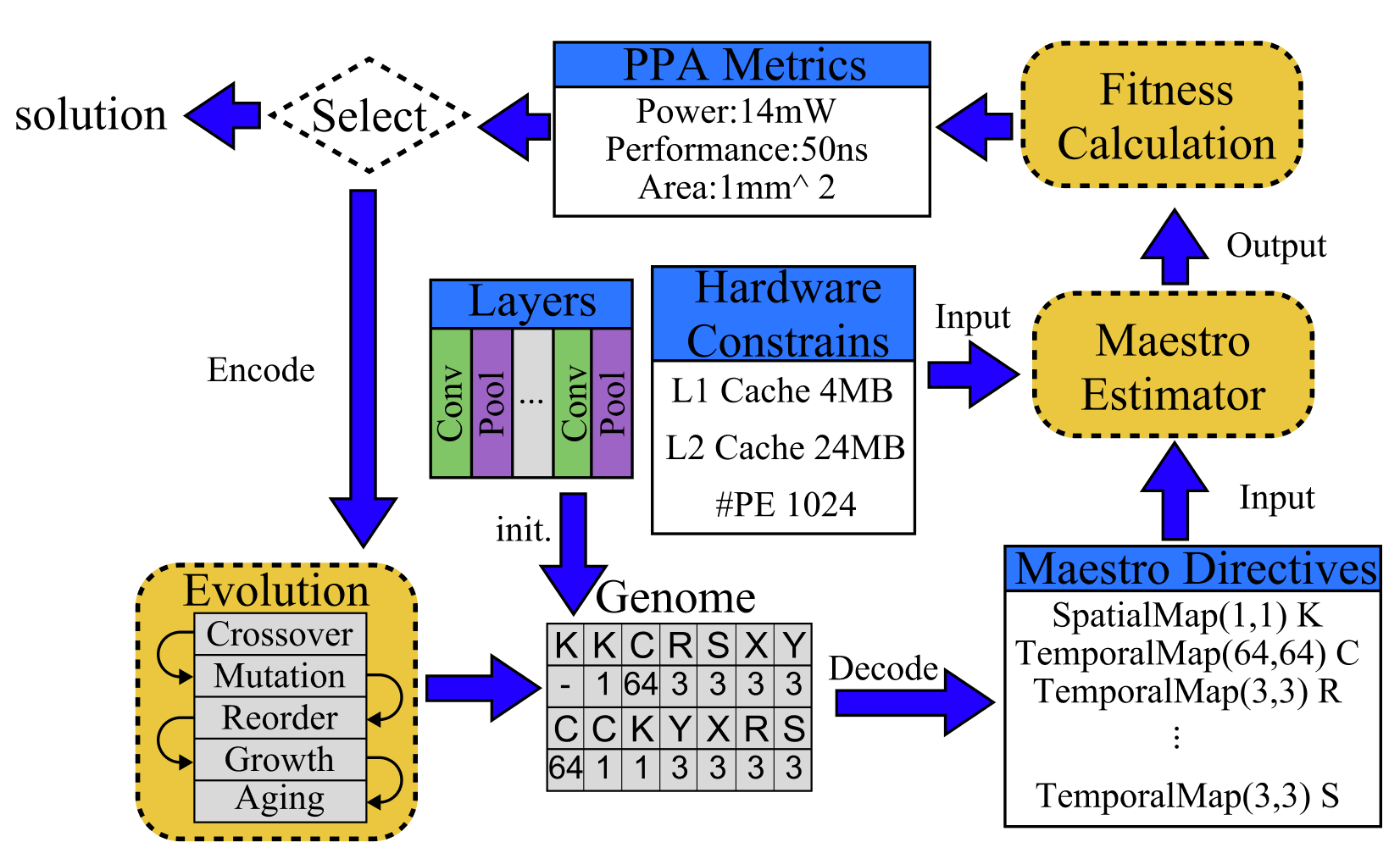}
    \caption{Chip mapping workflow. }
    \label{fig:hw-solver-workflow}
    \vspace{-3mm}
\end{figure}

Let \(\implementation{\chipconfigSet}\) be the set of solver and hardware configurations, where each \(\implementation{\chipconfig}\in\implementation{\chipconfigSet}\) specifies quantities such as the number of \glsplpe and the sizes of the \glslone and the \glsltwo.
Let \(M\) be the set of legal mappings, where each $m \in M$ is a list of MAESTRO~\cite{MAESTRO} directives, as shown in \cref{fig:hw-solver-workflow}.
The set $\implementation{\chipconfigSet}$ contains the single configuration of \cref{tab:hw-config}, which follows the generic spatial accelerator of~\cite{gamma} with \glsplpe, \glslone and \glsltwo buffers, and a \glsnoc.
Its 1024 \glsplpe lie between the 168 of edge accelerators and the 65{,}536 of cloud accelerators~\cite{eyeriss,tpu}.
The \glslone and \glsltwo sizes are upper bounds from the cloud platform of~\cite{gamma}, and the \glsnoc and off-chip bandwidths are unconstrained.
Other configurations can be added to $\implementation{\chipconfigSet}$ to explore different design spaces.

Since the chip metrics have three components, the solver also needs a preferred metric $\implementation{o} \in \implementation{O}= \{\textsc{Latency}, \textsc{Energy}, \textsc{Area}\}$, as \cref{rmk:priority-as-imp} explains.
For an embedding configuration, chip configuration, preference, \glsga epoch number \(t\), and random outcome \(\omega\), the solver produces a mapping $\hat{m}_{t,\implementation{o}}(\resource{\ell},\implementation{\chipconfig},\omega) \in M$ with the chip metrics tuple
\begin{equation}
\label{eq:hw-ppa-rv}
    \mathtt{X}_{t,\implementation{o}}
    =
    \operatorname{Eval}\!
    \left(
        \hat{m}_{t,\implementation{o}}
        (\resource{\ell},\implementation{\chipconfig},\omega)
    \right)
    \in
    \mathbb{R}_{+}^{3},
\end{equation}
where \(\operatorname{Eval}(\cdot)\) is the MAESTRO evaluation.
The orange $\implementation{o} \in \implementation{O}$ denotes the preference, which is part of the implementation, and the uncolored $o \in O$ indexes \glslea tuples, as in~$[\cdot]_o$.
Following \cref{sec:probabilistic-framework}, the solver is modeled as
\begin{equation}\label{eq:Psi_solver_whole}
\Psi_{\mathrm{solver}}: \resource{L} \times \implementation{\chipconfigSet} \times \implementation{O} \times \mathbb{N}_+ \times \Omega_{\mathrm{solver}} \to M\to \mathbb{R}_+^3,
\end{equation}
which maps each embedding configuration, chip configuration, preference, number of epochs, and random outcome to a mapping and then to its \glslea metrics.
The sample space $\Omega_{\mathrm{solver}}$ includes the randomized initialization, mutation, crossover, and all other stochasticity of the search.
Each~$\resource{\ell}$, $\implementation{\chipconfig}$, $\implementation{o}$, and~$t$ induce a conditional distribution~$\inducedDist{\cdot}{\resource{\ell},\implementation{\chipconfig},\implementation{o},t}$ over the \glslea metrics in~$\mathbb{R}_+^3$, as in \cref{sec:surrogate-model}, and the probability of achieving the \functionality{\glslea metrics}~$\functionality{\mathtt{X}}$ is
\begin{equation}\label{eq:probability-eq-for-solver}
\begin{aligned}
\pi = \inducedDist{
\bigl\{
[\functionality{\mathtt{X}}']_{o} \mid [\functionality{\mathtt{X}}']_{o} \geq [\functionality{\mathtt{X}}]_{o},
\forall o \in O
\bigr\}
}{
\resource{\ell},
\implementation{\chipconfig},
\implementation{o},
t},
\end{aligned}
\end{equation}
where $[\mathtt{X}]_o$ is the value of $\mathtt{X}$ on metric $o$.

\subsubsection{Surrogate Model}
Assuming sufficient on-chip and off-chip bandwidth, the \glslea metrics of a network $\resource{\ell} = (\mathtt{layer}_1, \dots, \mathtt{layer}_N)$ are the sums of those of its layers, so it suffices to characterize the distribution of each layer:
\begin{equation}
    \mathtt{X}_{\mathrm{net},t} = \sum_{n=1}^{N} \mathtt{X}_{n,t},
    \qquad
    \mathtt{X}_{n,t}
    \sim
    \mathcal{S}_t(\mathrm{layer}_{n}, \implementation{\chipconfig},  \implementation{o}).
\label{eq:uncertainty-solver}\end{equation}
Unlike training, the \glsga solver can be run many times at each epoch~$t$, and the $R$ i.i.d.\ samples $\{\mathtt{X}^{(r)}_t\}_{r=1}^{R}$ directly characterize $\mathcal{S}_t(\mathrm{layer}, \implementation{\chipconfig},  \implementation{o})$.
We run $R = 200$ independent \glsga runs of $3{,}000$ epochs each and fit $\mathcal{S}_{t_i}(\mathrm{layer}, \implementation{\chipconfig},  \implementation{o})$ with a parameterized model at several checkpoints $t_i$, which gives parameter vectors $\boldsymbol{\vartheta}_\implementation{o}(t_i)$.
A regression of each component $\vartheta_\implementation{o} \in \boldsymbol{\vartheta_\implementation{o}}$ over the epochs then gives $\hat{\boldsymbol{\vartheta}}_\implementation{o}(t)$ for every epoch $t$.
Empirically, the support of $\mathcal{S}_{t}(\mathrm{layer}, \implementation{\chipconfig},  \implementation{o})$ is a one-dimensional manifold, so we model only the distribution of the prioritized metric $[\mathtt{X}_t^{(r)}]_\implementation{o}^{-1}$ and recover the other two metrics by an indexing method.
For a fixed solver budget $t$, the surrogate is the three-component truncated mixture
\begin{equation}
\begin{split}
    f_{[\mathtt{X}_t]_o^{-1}}([\mathtt{x}]_o^{-1} \mid \mathtt{layer})
    = & \; w_1 f_1([\mathtt{x}]_o^{-1}) + w_2 f_2([\mathtt{x}]_o^{-1}) +
    \\
    w_3 f_3([\mathtt{x}]_o^{-1}),
    & \text{    s.t. } w_1 + w_2 + w_3 = 1,\; w_\cdot \ge 0,
\end{split}
\label{eq:hw-mixture-pdf}
\end{equation}
where the \textbf{peak component} $f_1$ and the \textbf{tail component} $f_3$ are left-truncated normal distributions for the near-optimal mappings and for the poor mappings trapped in distant local optima.
The \textbf{middle component} $f_2$ is a shifted Gamma distribution for the mappings of moderate quality.
The parameterization, the maximum-likelihood fit at the checkpoints, the regression over the epochs, and the nearest-neighbor indexing are detailed in Appendix~\ref{app:chip-design}.
The probability of getting \glslea metrics better than a tuple $\mathtt{x}$ is then estimated by Monte Carlo sampling with $K$ samples:
\begin{equation}
    \hat{p}_\mathrm{map}\bigl(\mathtt{X}_{\mathrm{net},t} \geq \mathtt{x}\bigr)
    = \frac{1}{K}\sum_{k=1}^{K}
      \mathbf{1}\!\Biggl(\sum_{n=1}^{N} \mathtt{X}_{n,t}^{(k)}\ge \mathtt{x} \Biggr).
    \label{eq:hw-mc-prob}
\end{equation}

\subsubsection{\texorpdfstring{\glsmdpi}{MDPI} Model}
The implementation set of the \bluecircled{2}\textbf{chip design} block is \(\implementation{\chipconfigSet}\times\implementation{O}\times\mathbb{N}_{+}\) of chip configurations, metric priorities, and solver epochs.
Its resources are the embedding configuration \(\resource{\ell}\in\resource{L}\), the solver budget \(\resource{b}\in\resource{B}\), and confidence \(\resource{\pi^{-1}}\), and it provides the chip metrics \(\functionality{A^\star}\), \(\functionality{E^\star}\), and \(\functionality{L^\star}\) in $\mathbb{R}_+$, the reciprocals of the physical values.
With \(\resource{b}_{\mathrm{map}}(\implementation{\chipconfig},\implementation{o},t)\) the standardized computational budget of running the solver for \(t\) epochs, the \glsmdpi is
\begingroup\setlength{\multlinegap}{0pt}%
\begin{multline}\label{eq:chip-design-pipeline-mdpi}
    \maketup{\implementation{h}, \implementation{o}, t}
    \mapsto
    \Big\{
    \maketup{\maketup{\functionality{A^\star}, \functionality{E^\star}, \functionality{L^\star}},\maketup{\resource{\ell}, \resource{b}', \resource{\pi^{-1}}}} \mid
    \resource{b}' \geq \resource{b}_{\mathrm{map}}(\implementation{\chipconfig},\implementation{o},t),
    \\
    \resource{\pi} \leq \hat{p}_\mathrm{map}\bigl(\mathtt{X}_{\mathrm{net},t} \geq \maketup{\functionality{A^\star}, \functionality{E^\star}, \functionality{L^\star}};\, \resource{\ell}, \implementation{h}, \implementation{o}\bigr)
    \Big\}.
\end{multline}\endgroup
Since the Monte Carlo estimator is inexpensive with the per-layer surrogates, the \glsmdpi is populated by sweeping all implementations over a grid of $(t, \pi)$ values.
Because interconnections depend only on the interfaces, other models, such as new surrogates for the same solver or other solvers, can be added as new implementations.

\subsection{Fabrication}
\label{sec:fabrication}

\glsnnp{}s are fabricated on silicon wafers, and each wafer is cut into dies.
A larger die fits fewer times on a wafer and is more likely to contain one of the random defects that make a die fail, so the number of working chips falls quickly as the die area grows.
Randomness in the defects induces a random number of working chips for each wafer.
The designer can choose a process node, inducing a trade-off between the number of working chips and the monetary cost of fabrication.

The \bluecircled{3}\textbf{fabrication} block represents this trade-off:
it requires \resource{Chip Area} at the reference process node, fabrication \resource{Cost}, and \resource{Confidence}, while providing the number of \functionality{Yielded Chips} and their \functionality{Process Node} for migration.
We further consider the environmental impact of the fabrication process by incorporating \resource{Water} usage, showing the flexibility of the co-design framework to include different aspects of a design in one formulation.
Cost and water are both determined by the number of processed wafers, with only yield being random; the stochastic model below therefore involves only \resource{Chip Area} and \resource{Cost}, and \resource{Water} enters the \glsmdpi in \cref{eq:fabrication-pipeline}.

\subsubsection{Stochasticity of Chip Fabrication Process}
\begin{definition}[Fabrication of Chips as Budgeted Stochastic
\glsmdpi]
    The yielded number of chips is modeled by a map
    \begin{equation}\label{eq:psi_fab}
        \Psi_{\text{fab}}: \implementation{\fabconfigSet} \times \mathbb{R}_+ \times \mathbb{R}_+ \times\Omega_{\mathrm{fab}}\to \mathbb{N}
    \end{equation}
    from each fabrication technology $\implementation{\fabconfig} \in \implementation{\fabconfigSet}$, \resource{Chip Area} in $\mathbb{R}_+$, and \resource{Monetary Cost} in $\mathbb{R}_+$ to the \functionality{Yielded Chip Number}, where $\Omega_{\mathrm{fab}}$ is the randomness of the defects.
    Each~$\implementation{\fabconfig}$, \resource{Chip Area}~$\resource{A}$, and \resource{Monetary Cost}~$\resource{C}$ induce a conditional distribution~$\inducedDist{\cdot}{\implementation{\fabconfig},\resource{A},\resource{C}}$ over~$\mathbb{N}$, as in \cref{sec:surrogate-model}.
\end{definition} 
The probability of yielding at least \functionality{n} chips is
\begin{equation}\label{eq:possibility-fab}
    \pi = \inducedDist{\makeset{\functionality{n}' \mid \functionality{n}' \geq \functionality{n}}}{\implementation{\fabconfig},\resource{A},\resource{C}},
\end{equation}
and we approximate the distribution by $\mathcal{T}$ for numerical results:
\begin{equation}
    \functionality{n} \sim \inducedDist{\cdot}{\implementation{\fabconfig},\resource{A},\resource{C}}
    \approx
    \mathcal{T}(\implementation{\fabconfig}, \resource{A}, \resource{C}).
\label{eq:uncertainty-fab}\end{equation}

Since the stochasticity mainly arises from defects, we instantiate $\mathcal{T}$ with the negative binomial yield model, which captures defect clustering:
\begin{equation}
    Y(d_0, \alpha, \resource{A}) = \left(1 + \frac{d_0 \cdot \resource{A}}{\alpha}\right)^{-\alpha},
\end{equation}
where $d_0$ is the defect density in defects/cm$^2$ and $\alpha$ is the clustering parameter.
Both are uniformly distributed, $d_0 \sim \mathcal{U}(d_0^{\min}, d_0^{\max})$ and~$\alpha \sim \mathcal{U}(\alpha^{\min}, \alpha^{\max})$, over ranges that the technology~$\implementation{\fabconfig}$ sets in \cref{tab:process_params}.

\begin{figure}[t]
    \centering
    \includegraphics[width=0.3\textwidth]{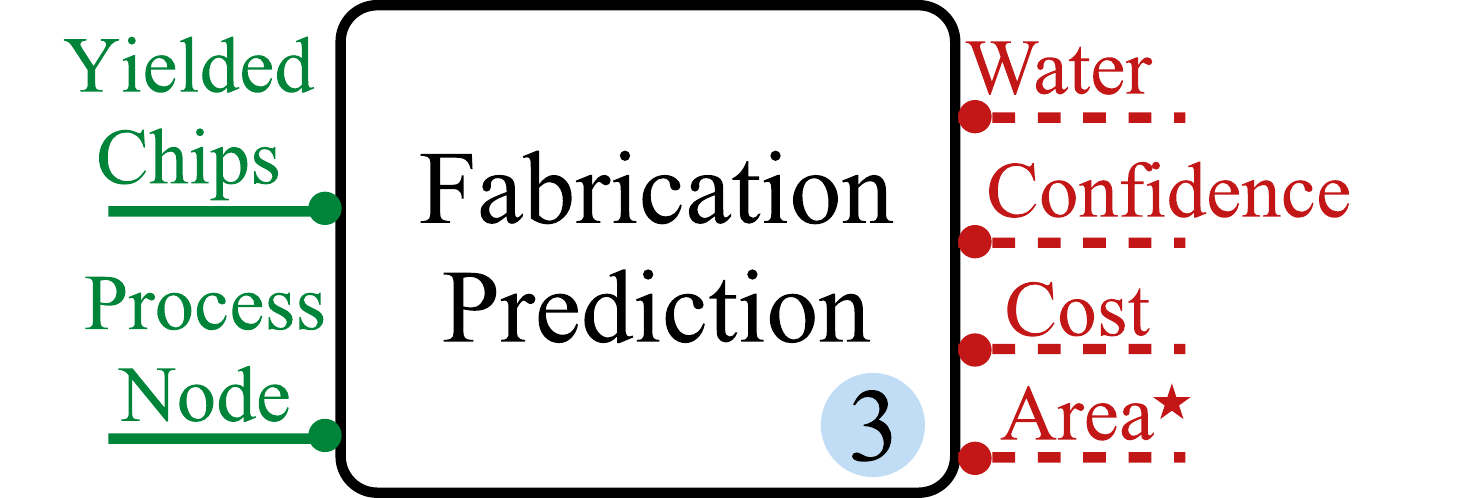}
    \caption{The \bluecircled{3}\textbf{Fabrication} \gls{abk:mdpi} block.
    }
    \label{fig:fab_mdpi}
\end{figure}

\begin{table}[t]
\centering
\caption{\centering Process node parameters \\for yield modeling and water use~\cite{boyd2009lca,bardon2020ppace}.}
\label{tab:process_params}
\setlength{\tabcolsep}{4pt}
\begin{tabular}{@{}lcccc@{}}
\toprule
\textbf{Process} & \textbf{Wafer} & $\mathbf{d_0}$ & $\mathbf{\alpha}$ & \textbf{Water} \\
\textbf{Node} & \textbf{Cost (\$)} & \textbf{Range} & \textbf{Range} & \textbf{(\unit{\litre}/wafer)} \\
\midrule
\qty{65}{\nano\metre} & 3,500 & $[0.10, 0.20]$ & $[1.0, 3.0]$ & 10,063 \\
\qty{45}{\nano\metre} & 4,200 & $[0.08, 0.15]$ & $[4.0, 8.0]$ & 5,316 \\
\qty{28}{\nano\metre} & 5,000 & $[0.03, 0.05]$ & $[2.0, 5.0]$ & 4,241 \\
\bottomrule
\end{tabular}
\vspace{-4mm}
\end{table}

With the dies-per-wafer estimate
\begin{equation}
    N_{\text{dpw}}(\resource{A}) = \left\lfloor \frac{\pi (D_w/2)^2}{\resource{A}} - \frac{\pi D_w}{\sqrt{2\resource{A}}} \right\rfloor,
\end{equation}
for the wafer diameter~$D_w$, which is \qty{300}{\milli\metre} in our experiments, a wafer has $N_{\text{good}} = \lfloor N_{\text{dpw}} \cdot Y \rfloor$ functional dies.
A budget~$\resource{C}$ buys $\left\lfloor \resource{C}/C_w \right\rfloor$ wafers of cost~$C_w$, which use $w \left\lfloor \resource{C}/C_w \right\rfloor$ liters of water, with $w$ the water per wafer in \cref{tab:process_params}.
Hence the \functionality{yielded chip number $n$} is
\begin{equation}\label{eq:n-fab-map}
    \functionality{n} = \left\lfloor \frac{\resource{C}}{C_w} \right\rfloor \cdot \left\lfloor N_{\text{dpw}}(\resource{A}) \cdot Y(d_0, \alpha, \resource{A}) \right\rfloor,
\end{equation}
and $\mathcal{T}(\implementation{\fabconfig}, \resource{A}, \resource{C})$ in \cref{eq:uncertainty-fab} is exactly the distribution of $\functionality{n}$ induced by those of $d_0$ and~$\alpha$.
We estimate it by Monte Carlo sampling: for each configuration, we draw $M$ samples of $(d_0, \alpha)$ and evaluate \cref{eq:n-fab-map}.

The probability $\pi$ in \cref{eq:possibility-fab} is then estimated empirically through
\begin{equation}
    F_\mathcal{T}\left(\functionality{n};\,\implementation{\fabconfig}, \resource{A}, \resource{C}\right) \approx \frac{1}{M}\sum_{m=1}^{M} \mathbf{1}[\functionality{n}^{(m)} \leq \functionality{n}],
\end{equation}
where $F_\mathcal{T}$ is the \gls{abk:cdf} of $\mathcal{T}$.

\subsubsection{\texorpdfstring{\glsmdpi}{MDPI} Model}

The fabrication implementation set is \(\implementation{\fabconfigSet}\), and the block provides \functionality{Yielded Chips} \(\functionality{n}\in\mathbb{N}\) and the \functionality{Process Node} $\functionality{\nu}$ of the selected implementation, and consumes  \resource{Chip Area} at the reference node, fabrication \resource{Cost}, \resource{Water}, and \resource{Confidence}. 
The realized \gls{abk:mdpi} is
\begin{equation}
\begin{aligned}
    \label{eq:fabrication-pipeline}
    \implementation{\fabconfig}
    &
    \mapsto
    \Big\{
    \maketup{\maketup{\functionality{n}, \functionality{\nu}},\maketup{\resource{A^{\star}}, \resource{C}, \resource{W}, \resource{\pi^{-1}}}} \mid
    \functionality{\nu} = \nu(\implementation{\fabconfig}),
    \\
    &\qquad
    \resource{W} \geq w(\implementation{\fabconfig}) \left\lfloor \resource{C}/C_w(\implementation{\fabconfig}) \right\rfloor,
    \\
    &\qquad
    \resource{\pi} \leq 1-F_\mathcal{T}\left(\functionality{n};\,\implementation{\fabconfig}, m_A(\nu(\implementation{\fabconfig}))\,\resource{A^{\star}}, \resource{C}\right)
    \Big\},
\end{aligned}
\end{equation}
where $\resource{A^\star}$ is the \resource{Chip Area} at the reference node, $\nu(\implementation{\fabconfig})$, $C_w(\implementation{\fabconfig})$, and $w(\implementation{\fabconfig})$ are the node, the wafer price, and the water per wafer of $\implementation{\fabconfig}$ in \cref{tab:process_params}, and $m_A(\nu(\implementation{\fabconfig}))\,\resource{A^\star}$ is the die area at this node, with the area factor $m_A$ of \cref{tab:migration}.

\subsection{Computation Distribution Planner}\label{sec:computation-planner}
Training candidate networks and exploring chip mappings need extensive compute on both CPUs and GPUs.
These CPUs and GPUs run the training and mapping solvers and are distinct from the chip being designed.
A machine with larger throughput may cost more and draws more power.
Consequently, there is a trade-off between the cost, time, and power consumption of the compute platform, captured by an \glsmdpi that allocates computation budgets to the \bluecircled{1}\textbf{network selection and training} block and the \bluecircled{2}\textbf{chip design} block.

\subsubsection{Abstract Model of Computation Distribution Planner}
Since workloads have different bottlenecks, the \bluecircled{4}\textbf{Planner} is decomposed into CPU and GPU paths, as shown in \cref{fig:Computation}.
Each path has a \textbf{calibration map} and \textbf{hardware alternatives}, and the calibration maps normalize the performance of the alternatives into unified computation budgets for training and mapping.
Let $\functionality{B}$ be the \glsposet of hardware-normalized \functionality{computational budgets}, $\resource{T}$ the \glsposet of \resource{time durations}, and $\resource{K}$ the \glsposet of hardware \resource{throughputs}.
The \emph{calibration map} is a monotone function $\mathrm{cal}: \resource{T} \times \resource{K} \to \functionality{B}$, indicating a more powerful machine and a longer time provide at least as much computation.
The time, power, and monetary cost of CPU, $\langle \resource{T_{\mathrm{cpu}}}, \resource{P_{\mathrm{cpu}}}, \resource{C_{\mathrm{cpu}}} \rangle$, and GPU, $\langle \resource{T_{\mathrm{gpu}}}, \resource{P_{\mathrm{gpu}}}, \resource{C_{\mathrm{gpu}}} \rangle$, aggregate as
$
    \resource{T_{\mathrm{total}}} = \resource{T_{\mathrm{cpu}}} + \resource{T_{\mathrm{gpu}}},
    \resource{P_{\mathrm{total}}} = \resource{P_{\mathrm{cpu}}} + \resource{P_{\mathrm{gpu}}},
    \resource{C_{\mathrm{total}}} = \resource{C_{\mathrm{cpu}}} + \resource{C_{\mathrm{gpu}}},
$
where the sum of times assumes sequential execution as the worst case.
Parallel execution would use $\resource{T_{\mathrm{total}}} = \max(\resource{T_{\mathrm{cpu}}}, \resource{T_{\mathrm{gpu}}})$, which is also monotone.

\subsubsection{\texorpdfstring{\glsmdpi}{MDPI} Model}
\begin{figure}[t]
    \centering
    \includegraphics[width=\columnwidth]{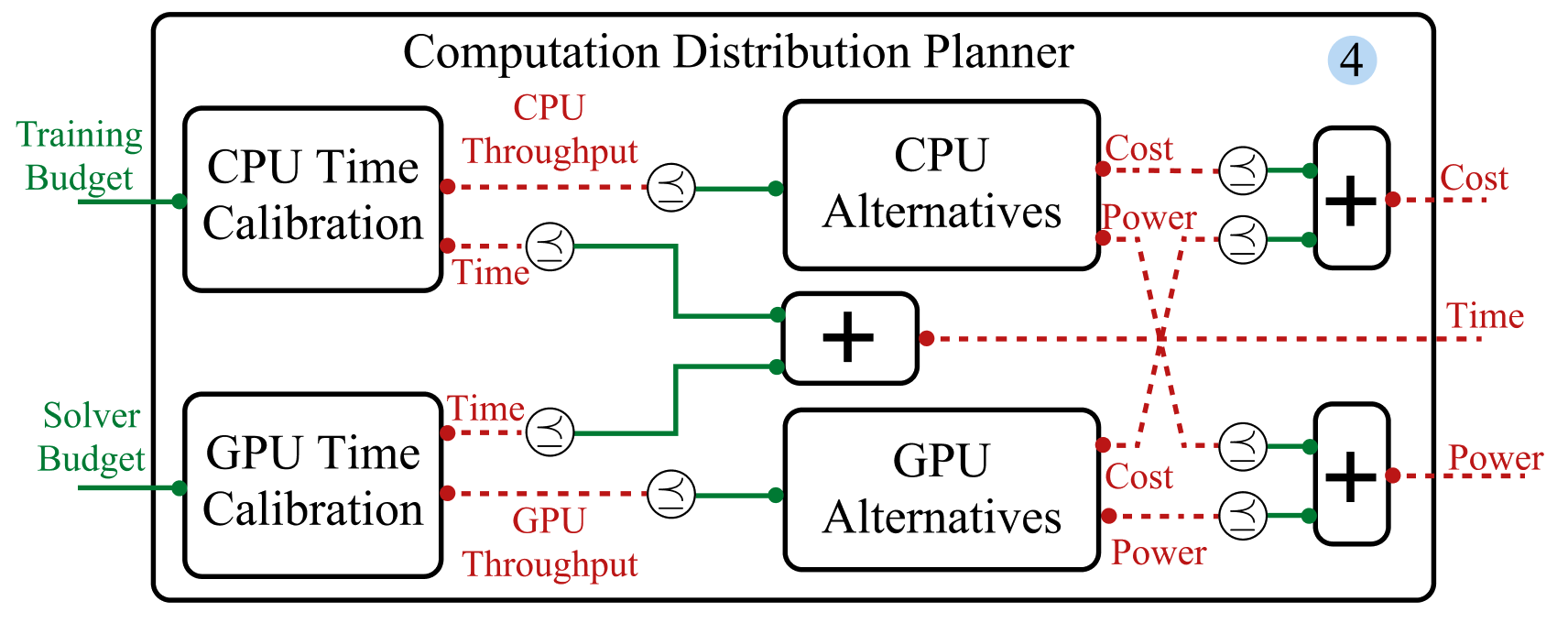}
    \caption{The \bluecircled{4}\textbf{Computation Distribution Planner} \gls{abk:mdpi} block. It decomposes computational requests into CPU and GPU workloads. This block illustrates the \gls{abk:mdpi} encapsulation property.}
    \label{fig:Computation}
    \vspace{-3mm}
\end{figure}

\Cref{fig:Computation} shows how the \textbf{calibration maps} and \textbf{hardware alternatives} form the planner \glsmdpi.
Its implementations are CPU and GPU choices that provide \functionality{Training Budget} \(\functionality{b}_{\mathrm{train}}\in\functionality{B}\) and \functionality{Hardware Mapping Solver Budget} \(\functionality{b}_{\mathrm{map}}\in\functionality{B}\) and consume \resource{Time}, \resource{Power}, and \resource{Monetary Cost}.

In general, $\functionality{B}$ could be a complicated \glsposet reflecting compute platform constraints, such as memory limiting the trainable network size or bandwidth limiting data flow.
Here, $\functionality{B} = \mathbb{R}_+$ is the wall-clock computing time on standard hardware, $\resource{K} = \mathbb{R}_+$ is the relative computing speed, and $\mathrm{cal}(t,\kappa) := t \times \kappa$.
With no design choice in the calibration maps, their \glsplmdpi have a single implementation:
\begin{equation}
\begin{aligned}
    \makeset{\implementation{\star}}
    &
    \to
    \dpOf{\functionality{B}}{\resource{K} \times \mathbb{R}_+}
    \\
    \implementation{\star}
    &
    \mapsto
    \Big\{
    \maketup{\functionality{b}, \maketup{\resource{\kappa}, \resource{t}}} \mid
    \functionality{b} \leq \mathrm{cal}(\resource{t}, \resource{\kappa}) = \resource{t} \times \resource{\kappa}
    \Big\}.
\end{aligned}
\end{equation}
Each CPU/GPU alternative is an implementation $\implementation{i}$ with monetary cost $\resource{C}_\implementation{i}$, power $\resource{P}_{\implementation{i}}$, and throughput $\functionality{\kappa}_\implementation{i}$:
\begin{equation}
\begin{aligned}
\label{eq:mdpi-computation-alternative}
    \implementation{I}
    &
    \to
    \dpOf{\functionality{K}}{\mathbb{R}_+ \times \mathbb{R}_+}
    \\
    \implementation{i}
    &
    \mapsto
    \Big\{\maketup{\functionality{\kappa}, \maketup{\resource{C}, \resource{P}}} \mid
    \functionality{\kappa} \leq \functionality{\kappa}_\implementation{i}, \resource{C} \geq \resource{C}_\implementation{i}, \resource{P} \geq \resource{P}_\implementation{i}
    \Big\}.
\end{aligned}
\end{equation}
The throughputs in \cref{tab:HW catalog} are measured on a reference workload and normalized against the Intel Core i7-12700KF for CPUs and the NVIDIA GeForce RTX 4070 Ti Super for GPUs.

\begin{table}[htbp]
\centering
\caption{Hardware catalog with normalized throughput.}
\begin{tabular}{llc}
\toprule
\textbf{Component} & \textbf{Type} & \textbf{Normalized Throughput $\kappa$} \\
\midrule
Intel Core i7-12700KF    & CPU & 1.00 (baseline) \\
Intel Core i5-13400F     & CPU & 0.64 \\
Intel Core i5-12600KF    & CPU & 0.72 \\
AMD Ryzen 5 7600X        & CPU & 0.80 \\
Intel Core i9-13900K     & CPU & 1.38 \\
AMD Ryzen 9 7950X        & CPU & 1.45 \\
\midrule
NVIDIA RTX 4070 Ti Super & GPU & 1.00 (baseline) \\
NVIDIA RTX 3080 Max-Q    & GPU & 0.62 \\
NVIDIA RTX 3070          & GPU & 0.78 \\
NVIDIA RTX A4000         & GPU & 0.81 \\
NVIDIA RTX 4090          & GPU & 1.82 \\
NVIDIA A100 80GB         & GPU & 2.95 \\
\bottomrule
\end{tabular}\label{tab:HW catalog}
\end{table}
\subsection{\texorpdfstring{Migration}{Migration}}\label{sec:migration}
An \glsnnp's  area, energy, and latency depend on the process node, yet the \bluecircled{2}\textbf{chip design} block evaluates them at one reference node.
We therefore consider the \bluecircled{5}\textbf{migration} block, which converts area, energy, and latency from the reference node to the node in which the \glsnnp is fabricated.
The Process Node is a \resource{resource} provided by \bluecircled{3}\textbf{fabrication}.

\subsubsection{\texorpdfstring{Abstract model}{Abstract model}}

Let $\resource{\mathcal{V}}$ be the poset of \resource{process nodes} with the discrete order, and $\mathbb{R}_+$ be the poset of chip areas, energies, and latencies.
The \emph{migration map} is then a monotone function:
\begin{equation}
\label{eq:psi_mig}
\mathrm{mig}: \resource{\mathcal{V}} \times \resource{\mathbb{R}_+} \times \resource{\mathbb{R}_+} \times \resource{\mathbb{R}_+} \to \functionality{\mathbb{R}_+} \times \functionality{\mathbb{R}_+} \times \functionality{\mathbb{R}_+},
\end{equation}
mapping each \resource{Process Node} $\resource{\nu}$ and the \resource{Chip Area}, \resource{Chip Energy}, and \resource{Chip Latency} at the reference node $\nu_0 \in \mathcal{V}$ to the \functionality{Chip Area}, \functionality{Chip Energy}, and \functionality{Chip Latency} at $\resource{\nu}$.
Monotonicity guarantees that, for any $\resource{\nu} \posetleq_{\resource{\mathcal{V}}} \resource{\nu}'$ and $\maketup{\resource{A^\star}, \resource{E^\star}, \resource{L^\star}} \leq \maketup{\resource{A^{\star\prime}}, \resource{E^{\star\prime}}, \resource{L^{\star\prime}}}$, $\mathrm{mig}(\resource{\nu}, \resource{A^\star}, \resource{E^\star}, \resource{L^\star}) \leq \mathrm{mig}(\resource{\nu}', \resource{A^{\star\prime}}, \resource{E^{\star\prime}}, \resource{L^{\star\prime}})$.
Intuitively, if one chip is smaller, more energy-efficient, or faster than another at the reference node, it remains so when both are fabricated in the same node.

\subsubsection{\texorpdfstring{\gls{abk:mdpi} model}{MDPI model}}

\begin{figure}[t]
    \centering
    \includegraphics[width=0.3\textwidth]{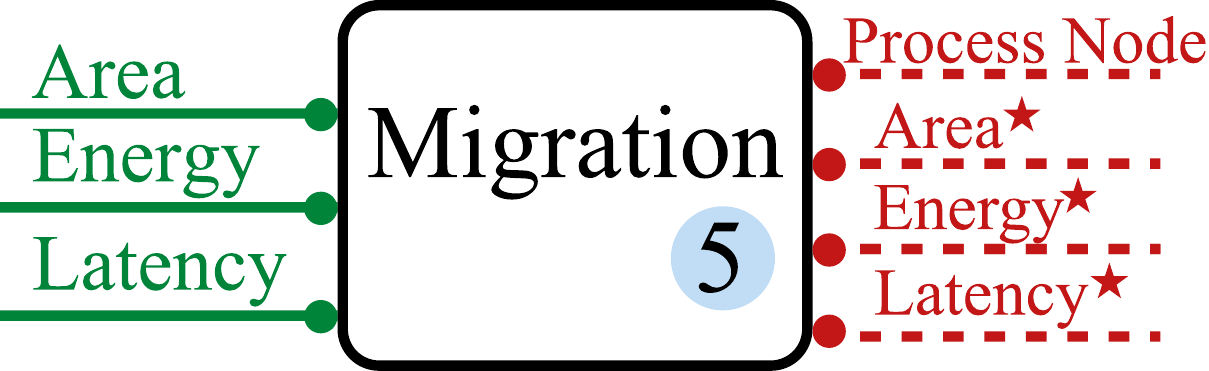}
    \caption{The \bluecircled{5}\textbf{Migration} \gls{abk:mdpi} block.
    }
    \label{fig:migration_mdpi}
    \vspace{-5mm}
\end{figure}

As shown in \cref{fig:migration_mdpi}, the block provides \functionality{Chip Area} $\functionality{A}$, \functionality{Chip Energy} $\functionality{E}$, and \functionality{Chip Latency} $\functionality{L}$ in $\mathbb{R}_+$ at node $\resource{\nu}$, while consuming the \resource{Process Node} $\resource{\nu} \in \resource{\mathcal{V}}$ and the \resource{Chip Area} $\resource{A^\star}$, \resource{Chip Energy} $\resource{E^\star}$, and \resource{Chip Latency} $\resource{L^\star}$ at the reference node.
Let $m_A(\nu), m_E(\nu), m_L(\nu) > 0$ be the area, energy, and latency at node $\nu$ relative to $\nu_0$.
Since $\resource{A^\star}$, $\resource{E^\star}$, and $\resource{L^\star}$ are reciprocals, the migration map is
\begin{equation}
\label{eq:migration-map}
    \mathrm{mig}(\resource{\nu}, \resource{A^\star}, \resource{E^\star}, \resource{L^\star})
    =
    \left\langle \frac{\resource{A^\star}}{m_A(\resource{\nu})}, \frac{\resource{E^\star}}{m_E(\resource{\nu})}, \frac{\resource{L^\star}}{m_L(\resource{\nu})} \right\rangle.
\end{equation}
Because we have no design choice in the migration map, this \glsmdpi has a single implementation, the migration rule $\implementation{\lambda}$, mapping to one specific \glsdp:
\begin{equation}
\label{eq:migration-mdpi}
\begin{aligned}
    \makeset{\implementation{\lambda}}
    &\to
    \dpOf{\mathbb{R}_+ \times \mathbb{R}_+ \times \mathbb{R}_+}{\resource{\mathcal{V}} \times \mathbb{R}_+ \times \mathbb{R}_+ \times \mathbb{R}_+}
    \\
    \implementation{\lambda}
    &\mapsto
    \Big\{
    \maketup{\maketup{\functionality{A}, \functionality{E}, \functionality{L}},\maketup{\resource{\nu}, \resource{A^\star}, \resource{E^\star}, \resource{L^\star}}} \mid
    \\
    &\qquad
    \maketup{\functionality{A}, \functionality{E}, \functionality{L}} \leq \mathrm{mig}(\resource{\nu}, \resource{A^\star}, \resource{E^\star}, \resource{L^\star})
    \Big\}.
\end{aligned}
\end{equation}

In this work, $\mathcal{V}$ contains the \qtylist{65;45;28}{\nano\metre} nodes.
We take \qty{45}{\nano\metre} as the reference node $\nu_0$, the node of the energy per arithmetic operation in our cost model~\cite{horowitz2014energy}, and use the scaling factors of DeepScaleTool~\cite{deepscaletool}, listed in \cref{tab:migration}.

\begin{table}[htbp]
\centering
\caption{Scaling factors of the migration block.}
\label{tab:migration}
\begin{tabular}{lccc}
\toprule
\textbf{Process Node} & \textbf{Area} $m_A$ & \textbf{Energy} $m_E$ & \textbf{Latency} $m_L$ \\
\midrule
\qty{65}{\nano\metre} & 2.04 & 1.59 & 1.23 \\
\qty{45}{\nano\metre} & 1 & 1 & 1 \\
\qty{28}{\nano\metre} & 0.35 & 0.59 & 0.83 \\
\bottomrule
\end{tabular}
\end{table}

\section{Simulation Results}\label{sec:simulation results}
The three case studies below validate the framework against the three challenges
identified in~\cref{sec:intro}.
Case Study~I demonstrates \emph{end-to-end co-design across abstraction levels}
by solving a concrete \glsnnp design problem and recovering Pareto-optimal
implementations spanning network training through wafer fabrication.
Case Study~II shows the ability of this framework to design \glsnnp for multiple scenarios. This case study also validates \emph{uncertainty quantification} by confirming that the
surrogate distributions faithfully capture observed stochasticity in all three
probabilistic blocks, and by demonstrating that \resource{Confidence}~$\resource{\pi^{-1}}$
functions as a first-class, continuously tunable design resource.
Case Study~III establishes \emph{algorithm–framework decoupling} by showing
that enriching the implementation set of a single block automatically improves
the composed Pareto front without modifying the surrounding co-design diagram.
The offline surrogate architecture used in all three studies also mitigates the \emph{evaluation-cost} challenge: with the surrogate models in place, each query runs in seconds without re-invoking the underlying solvers.
The downside is that the reliability of final results depends on quality of the surrogate models.
Pareto-optimal resource consumptions with corresponding implementations are computed as query results (\cref{def:mdpi-queries}) with the compositional solver \cite{censi2015mathematical,zardiniCoDesignComplexSystems2023}.

\subsection{Case Study I: End-to-End Co-Design of a Baseline \glsnnp}
\textbf{Scenario 0 (Baseline)} instantiates the framework for a first-generation
indoor cleaning robot whose primary objective is to minimize \resource{Cost}.
All five \functionality{functionalities} are set to 0, meaning that no limit is placed on chip \functionality{area}, \functionality{energy}, or \functionality{latency}, and no minimum \functionality{accuracy} or \functionality{yielded chip number} is required.
This scenario provides a reference operating point for the more demanding scenarios that follow.

\begin{figure}[tb]
  \includegraphics[width=\columnwidth]{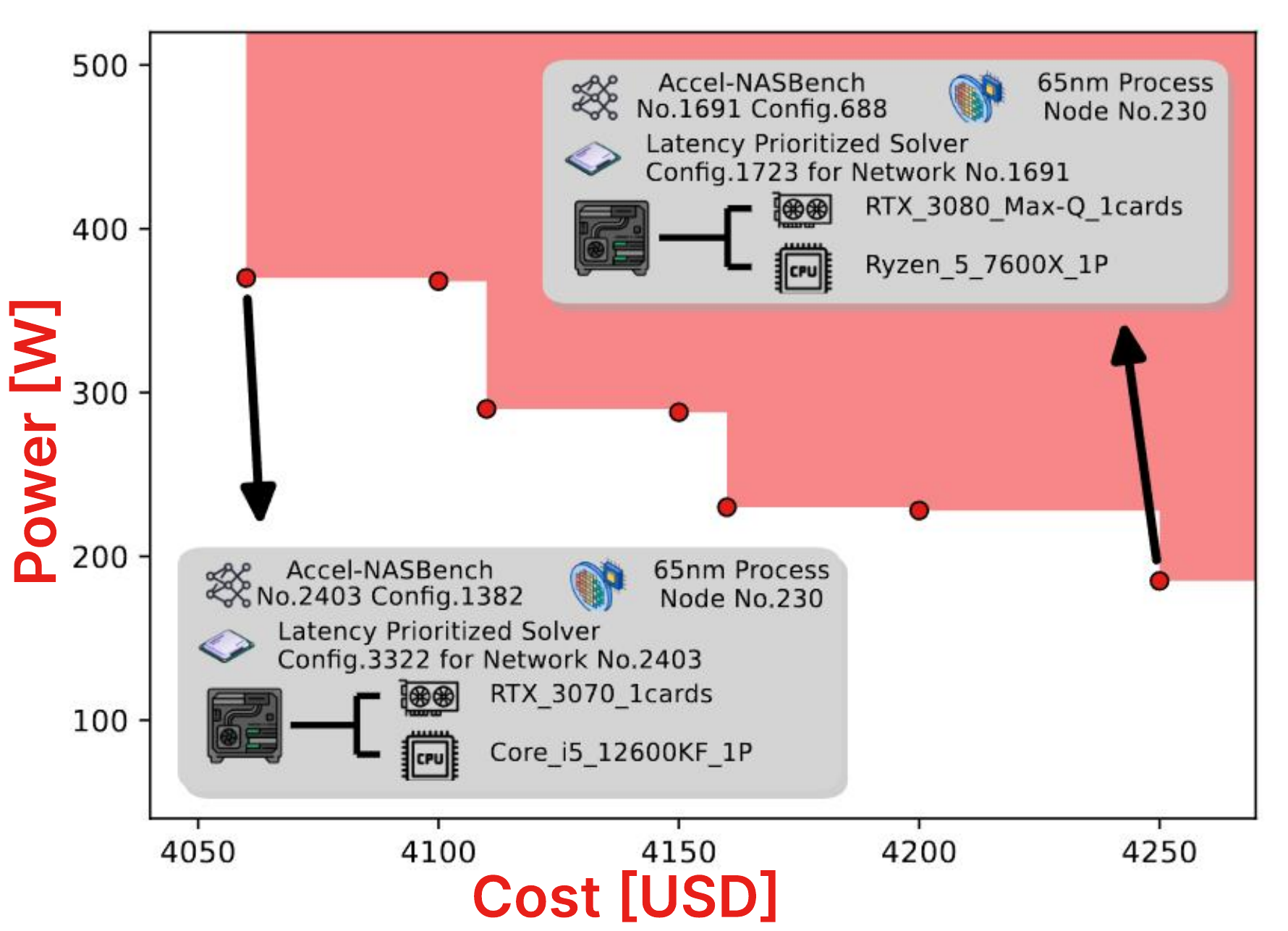}
  \caption{%
    \resource{Power}–\resource{Cost} Pareto front for \textbf{Scenario~0}.
    Two Pareto-optimal \implementation{implementations} are expanded to show the
    full design decisions selected by the framework across all five MDPIs.}
  \label{fig:pareto boundary}
\end{figure}

\Cref{fig:pareto boundary} presents the \resource{Power}–\resource{Cost} Pareto
front returned by the solver.
The front spans a compact range, i.e., \qtyrange{185}{370}{\watt} and \$4{,}060 to \$4{,}250, and reveals two qualitatively distinct regimes: a \emph{power-limited} region (upper-left) in which aggressive hardware choices reduce \resource{Cost} at the expense of \resource{Power}, and a \emph{cost-limited} region (lower-right) in which conservative choices sacrifice cost efficiency to stay within a tighter power envelope.

Two implementations are expanded in the figure to illustrate how the framework maps Pareto points to actionable design decisions.
The cost-optimal solution (\resource{Cost} $\approx$ \$4{,}060, \resource{Power} $\approx$ \qty{370}{\watt})
selects \implementation{Accel-NASBench Network No.\,2403 with Config.\,1382}, \implementation{a Latency-Prioritized Solver Config.\,3322}, and \implementation{\qty{65}{\nano\metre} Process Node No.\,230} for fabricating the \glsnnp{}, while network training and hardware mapping run on \implementation{one RTX\,3070 GPU} and \implementation{one Core~i5-12600KF CPU}.
No single block was optimized in isolation; the co-design loop jointly determines network architecture, mapping strategy, fabrication node, and compute hardware, a result that would be difficult to recover through sequential, block-by-block optimization.

\subsection{Case Study II: Multi-Scenario Evaluation with Distributional Uncertainty}
\subsubsection{Impact of Tightening Functionality Constraints}
To probe the framework's sensitivity to heterogeneous application requirements, we evaluate four scenarios (\Cref{tab:scenarios}) with progressively stricter \functionality{functionality} constraints.

\textbf{Scenario 1 (Area-constrained):}
Targets a compact production-line inspection robot operating in confined industrial
spaces. 
Only chip \functionality{area} is bounded ($\leq \qty{1.8}{\square\milli\metre}$);
\functionality{energy} and \functionality{latency} are unconstrained.

\textbf{Scenario 2 (Area- and Energy-constrained):}
Targets a smart hearing aid where both silicon footprint and battery life are
critical. 
The \functionality{area} limit is carried over from Scenario~1 and
\functionality{energy} is additionally capped at \qty{900}{\micro\joule} per inference.

\textbf{Scenario 3 (\glslea-constrained):}
Targets a neural implant for brain–computer interfaces, simultaneously tightening
all five \functionality{functionalities}: \functionality{area}
($\leq \qty{1.5}{\square\milli\metre}$), \functionality{energy}
($\leq \qty{900}{\micro\joule}$), \functionality{latency}
($\leq 10{,}205$ cycles), minimum \functionality{chip yield} ($\geq 10{,}000$
parts), and inference \functionality{accuracy} ($\geq 88\%$).
This scenario exercises the full coupling across all five MDPIs and leaves the
least slack in the feasible implementation set.

\begin{figure}[tb]
  \centering
  \includegraphics[width=0.45\textwidth]{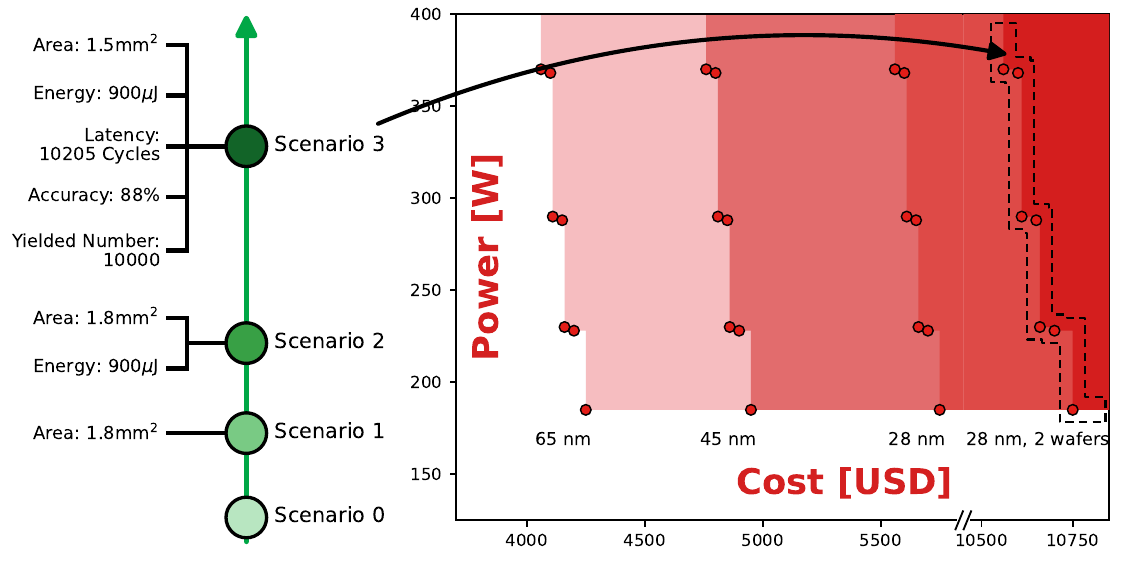}
  \caption{%
    \resource{Power}–\resource{Cost} Pareto boundaries for Scenarios 0–3,
    reflecting the compounding expenditure of multi-dimensional constraint
    satisfaction.}
  \label{fig:Compare scenarios}
  \vspace{-3mm}
\end{figure}

\begin{table}[htbp]
  \centering
  \renewcommand{\arraystretch}{1.2}
  \setlength{\tabcolsep}{4pt}
  \newcommand{\mcell}[1]{\begin{tabular}[c]{@{}c@{}}#1\end{tabular}}
  \begin{threeparttable}
    \caption{Functionality Constraints by Scenario}
    \label{tab:scenarios}
    \begin{tabular}{@{} l c c c c c @{}}
      \hline
      & \multicolumn{5}{c}{\textbf{\functionality{Functionality}}\tnote{1}} \\
      \cline{2-6}
      \textbf{Scenario} & \mcell{\functionality{Area} \\ (\unit{\square\milli\metre})} & \mcell{\functionality{Energy} \\ (\unit{\micro\joule})}
                        & \mcell{\functionality{Latency} \\ (cycles)} & \mcell{\functionality{Yield} \\ (chips)} & \functionality{Accuracy} \\
      \hline
      0 (Baseline)       & --  & --  & --  & --  & --  \\
      1 (Area)           & $\leq 1.8$ & -- & -- & -- & -- \\
      2 (Area \& Energy) & $\leq 1.8$ & $\leq 900$ & -- & -- & -- \\
      3 (\glslea)        & $\leq 1.5$ & $\leq 900$ & $\leq 10{,}205$ & $\geq 10{,}000$ & $\geq 88\%$ \\
      \hline
    \end{tabular}
    \begin{tablenotes}
      \footnotesize
      \item[1] \textit{\functionality{Functionalities} marked -- are} \textit{set to 0, i.e., not constrained.}
    \end{tablenotes}
  \end{threeparttable}
  \vspace{-3mm}
\end{table}

\Cref{fig:Compare scenarios} confirms a monotonic relationship between constraint tightness and resource expenditure: the Pareto boundary shifts outward with each successive scenario as the feasible implementation set shrinks and the solver must
draw on more compute, cost, or time to satisfy all \functionality{functionalities}.
The shift from \textbf{Scenario~2} to \textbf{Scenario~3} is the most pronounced, consistent with the five simultaneous constraints of the latter leaving negligible slack for the mapping and training solvers to exploit.
This monotonic progression is a direct consequence of the order-theoretic structure of the MDPI formalism, and not an artifact of the specific solvers used.

\subsubsection{Confidence as a Continuously Tunable Design Knob}
\label{sec:confidence_knob}
A key limitation of deterministic co-design frameworks is that they reduce probabilistic outcomes, i.e., training convergence, mapping quality, fabrication
yield, to point estimates, conflating the \emph{best-case} and \emph{expected-case} behavior of a design.
The proposed framework instead treats \resource{Confidence} $\resource{\pi^{-1}}$ as an explicit resource: the inverse success probability of jointly satisfying all stochastic \functionality{functionality} requirements.
This reframing converts a binary feasibility question into a continuous trade-off that the designer can navigate in the same Pareto framework used for
\resource{Time}, \resource{Power}, and \resource{Cost}.

\begin{figure}[tb]
  \centering
  \includegraphics[width=0.4\textwidth]{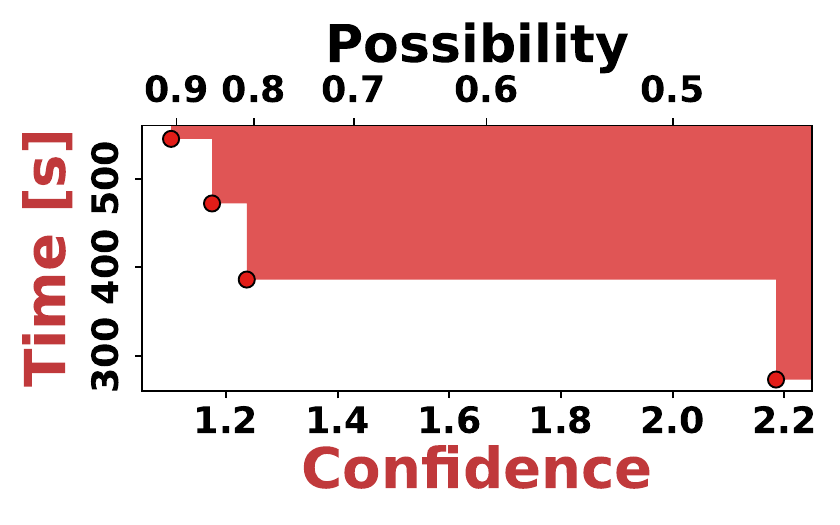}
  \caption{%
    \resource{Time}–\resource{Confidence} Pareto front for \textbf{Scenario~3}.
    The top axis reports the corresponding success probability.
    Each red marker is a Pareto-optimal \implementation{implementation}; the
    staircase profile reflects the discrete hardware and network catalogs.}
  \label{fig:possibility_pareto}
  \vspace{-3mm}
\end{figure}

\Cref{fig:possibility_pareto} shows the \resource{Time}–\resource{Confidence} Pareto front for the most constrained case, \textbf{Scenario~3}.
The front exhibits two distinct regimes separated by a sharp transition.

In the \emph{high-confidence regime} (left of the transition), each marginal increase in success probability requires a disproportionate investment in
\resource{Time}: both the training and mapping solvers must run substantially longer to push the probability mass of attainable \functionality{functionalities} above the required threshold.
This steep slope arises because Scenario~3's tight constraints leave little slack in the feasible implementation set, so the tail of the performance distribution must be reliably avoided rather than merely unlikely.

In the \emph{low-confidence regime} (right of the transition), the front drops sharply to much lower \resource{Time}: accepting a weaker probabilistic guarantee allows the solver to terminate early and still satisfy the \functionality{functionality} requirements on most but not all runs.
The magnitude of the gap between regimes is a direct, quantitative measure of the cost of reliability for this application—information that is entirely hidden
in deterministic formulations.

From a workflow perspective, \resource{Confidence} behaves as a knob whose optimal setting is application-dependent: a safety-critical neural implant (\textbf{Scenario~3}) warrants a high-confidence, resource intensive operating point, whereas an indoor cleaning robot (\textbf{Scenario~0}) may tolerate a looser guarantee
in exchange for substantially reduced compute budget.
The framework surfaces this trade-off explicitly, enabling principled rather than ad~hoc reliability budgeting.

\subsubsection{Validation of Surrogate Distributions Across All Three
               Probabilistic Blocks}
\label{sec:dist_fit}

Reliability of query results depends on the quality of the surrogate models.
We validate each of them.

\begin{figure*}[t]
  \centering
  \captionsetup[subfigure]{labelformat=simple}%
  \renewcommand{\thesubfigure}{(\alph{subfigure})}%
  \begin{subfigure}[t]{0.355\textwidth}
    \centering
    \includegraphics[width=\linewidth]{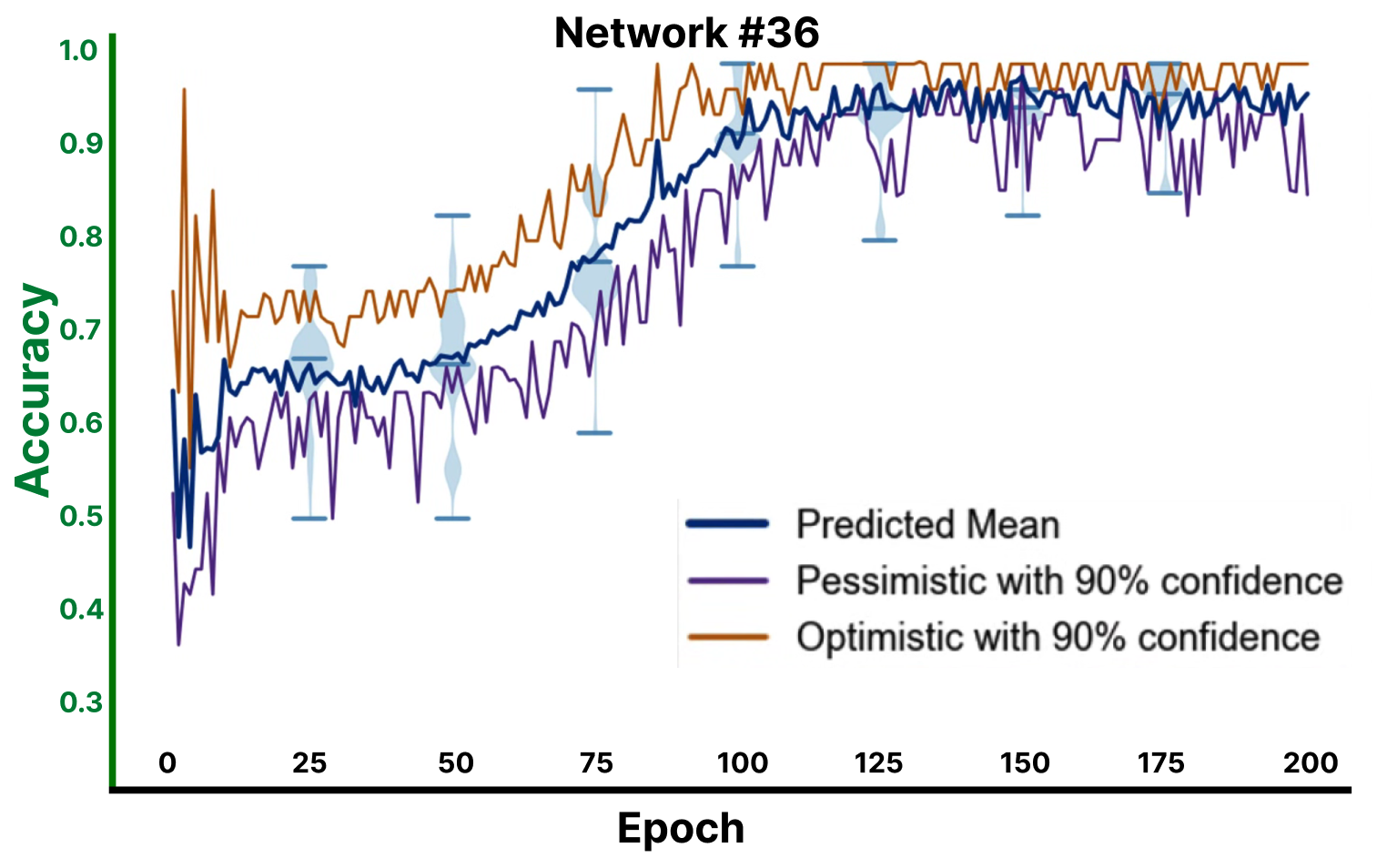}
    \subcaption{}\label{fig:dist_training}
  \end{subfigure}\hfill
  \begin{subfigure}[t]{0.277\textwidth}
    \centering
    \includegraphics[width=\linewidth]{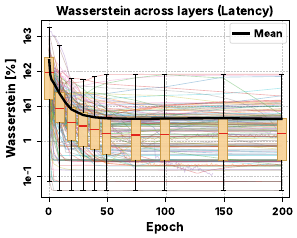}
    \subcaption{}\label{fig:dist_chip}
  \end{subfigure}\hfill
  \begin{subfigure}[t]{0.328\textwidth}
    \centering
    \includegraphics[width=\linewidth]{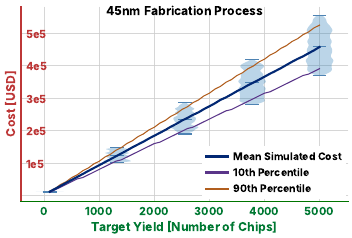}
    \subcaption{}\label{fig:dist_fabrication}
  \end{subfigure}
  \caption{%
    Surrogate distributions of the three probabilistic blocks.
    (a)~Fitted learning-curve surrogate for the \bluecircled{1}\textbf{Network Training
    Pipeline} (Network~\#36).
    (b)~Wasserstein-1 distance between the fitted mixture distribution
    (\cref{eq:hw-mixture-pdf}) and the empirical LEA distribution for the profiled network layers from HW-NAS-Bench and Accel-NASBench (coloured curves), as a function of solver budget
    (epochs); box plots show the spread across layers.
    (c)~\bluecircled{3}\textbf{Fabrication} block (45\,nm process node):
    manufacturing \resource{Cost} vs.\ target yield.
    }
  \label{fig:dist_fit}
  \vspace{-5mm}
\end{figure*}

\paragraph{\textbf{\bluecircled{1} Network Training Pipeline}}
\Cref{fig:dist_training} shows the fitted mean trajectory~$\mu(t\mid\theta)$ alongside the empirical 10th and 90th percentile accuracy curves for a representative network.
The parametric form from \cref{sec:nn-architectures} tracks both the central tendency and the spread of the empirical trajectories.

\paragraph{\textbf{\bluecircled{2} Chip Design Pipeline}}
We assess fit quality via the Wasserstein-1 distance between the fitted three-component mixture according to \cref{eq:hw-mixture-pdf}, and the empirical distribution of  in \Cref{fig:dist_chip}.
The Wasserstein distance decays  with solver budget and concentrates across network layers.

\paragraph{\textbf{\bluecircled{3} Fabrication Block}}
\Cref{fig:dist_fabrication} plots the mean simulated manufacturing cost and the 10th/90th percentile band as a function of target yield for the 45\,nm process
node.
The widening band at larger production volumes reflects the compounding effect
of per-wafer yield variance.

\subsection{Case Study III: Algorithm--Framework Decoupling}

A co-design framework that tightly couples the global objective to a specific
solver is fragile: improving or replacing any internal algorithm requires
reworking the entire exploration pipeline.
The MDPI formalism avoids this by making each block's external interface—its
\functionality{functionality} and \resource{resource} posets—the \emph{only}
contract visible to the rest of the diagram.
The internal algorithm that populates the implementation set is entirely opaque
to adjacent blocks, so it can be replaced without structural modifications to the
co-design diagram.

\subsubsection{Experimental Setup}

We validate this property on the \bluecircled{2}\textbf{Chip Design Pipeline} by adding to its implementation set the implementations of a second mapping solver, Stream~\cite{stream}, which evaluates each layer with the ZigZag cost model~\cite{zigzag}.

\subsubsection{Results}

\begin{figure}[t]
  \centering
  \includegraphics[width=0.4\textwidth]{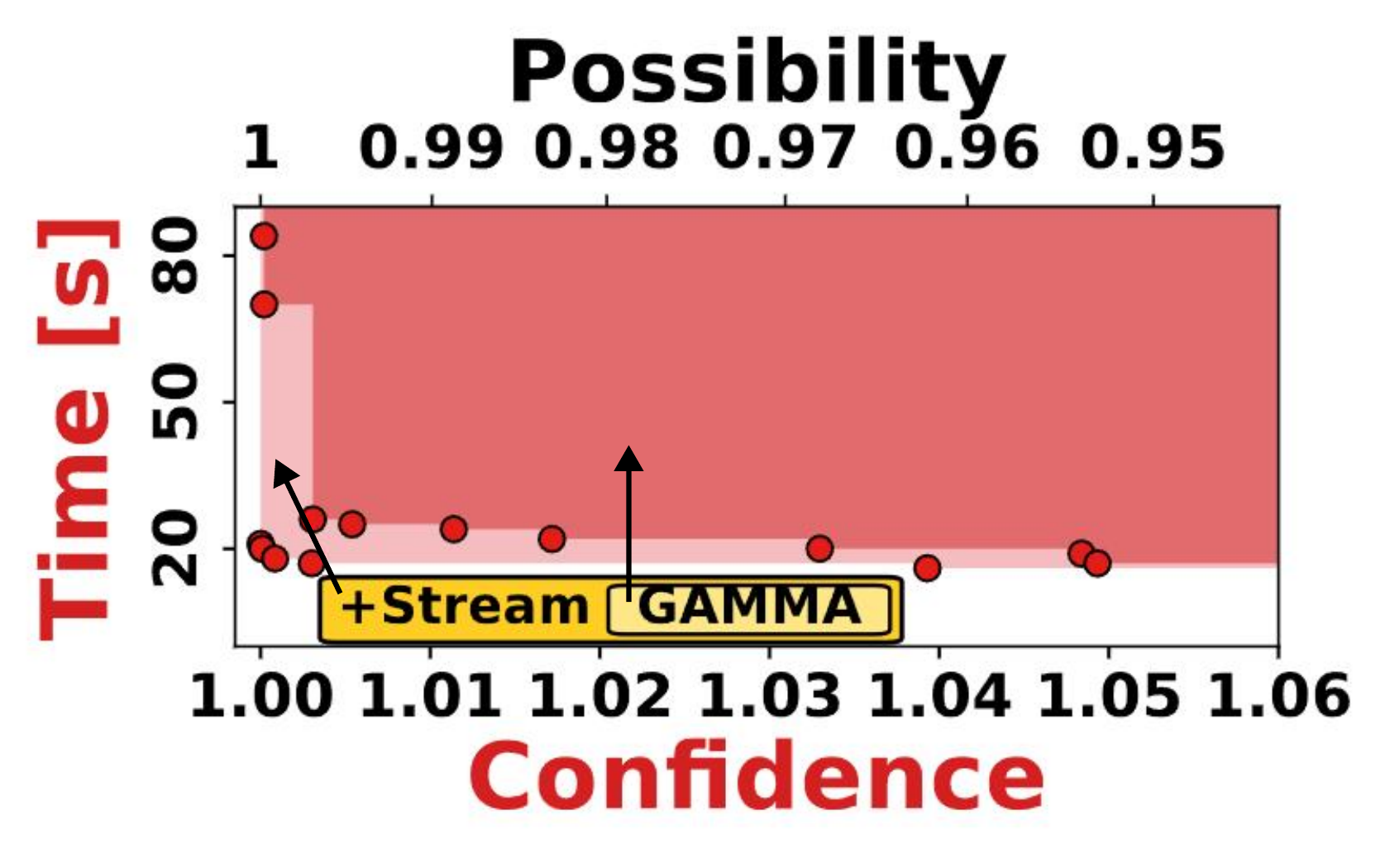}
  \caption{%
    Composed \resource{Time}–\resource{Confidence} Pareto fronts without and with the Stream implementations in the
    \bluecircled{2}\textbf{Chip Design Pipeline}.
}
  \label{fig:case study ii diff algos}
  \vspace{-4mm}
\end{figure}

We pose the baseline query of Scenario~0 twice, with the GAMMA implementations only and with the Stream implementations added, and compare the composed \resource{Time}--\resource{Confidence} fronts in \Cref{fig:case study ii diff algos}.
With Stream added, every design on the front uses a Stream implementation, and the front moves to a shorter \resource{Time} at every \resource{Confidence}.
The framework detects and propagates these improvements automatically: no block
other than \bluecircled{2}\textbf{Chip Design} is touched, and the surrounding co-design diagram is
structurally unchanged.

This behavior is a direct consequence of the MDPI composition rules: the global
optimization queries only the implementation set's Pareto-minimal elements, and
those elements change only when a superior implementation enters the set.
In practice, this means that researchers can independently improve any
block, replacing a training surrogate, upgrading the mapping solver, or refining
a yield model, and immediately observe the effect on the system-level Pareto
front, without redesigning the co-design methodology that connects them.

\section{Conclusion}
\label{sec:conclusion}
This paper introduced a monotone co-design framework that unifies neural network
selection and training, chip mapping, wafer-level fabrication, migration, and compute
resource allocation into a single compositional optimization problem.
The central design choice, modeling each subsystem as a monotone design problem
with implementation (\gls{abk:mdpi}) whose \functionality{functionality}–\resource{resource}
interface is decoupled from its internal algorithm, yields four properties that
prior hardware-aware NAS and co-exploration approaches do not simultaneously
provide: composition across abstraction levels, first-class uncertainty
quantification, offline-surrogate-based evaluation, and algorithm–framework
independence.

The three case studies translate these properties into concrete evidence.
Case Study~I demonstrated that the composed framework recovers Pareto-optimal
\glsnnp designs spanning network architecture through fabrication node,
producing actionable implementation choices that no single-level optimizer
could identify in isolation.
Case Study~II established the most distinctive contribution of the work: by
treating \resource{Confidence} $\resource{\pi^{-1}}$ as an explicit,
order-theoretic resource rather than a post-hoc diagnostic, the framework
converts the question of reliability from a binary feasibility check into a
continuously tunable design knob.
Case Study~III showed that  expanding the implementation set of a single block
automatically and correctly propagates into an improved global Pareto front,
with no modification to the surrounding co-design diagram—a property that
fundamentally changes the economics of iterative \glsnnp development by allowing
algorithmic improvements to be contributed and composed independently.

Together, these results position our formalism as a viable foundation for
principled, end-to-end \glsnnp co-design that is simultaneously rigorous
enough to carry formal guarantees, flexible to accommodate heterogeneous
surrogate models, and modular enough to evolve with the algorithms it wraps.

\subsection{Outlook}

Several directions offer natural extensions.
\paragraph{Richer models and design choices}
Reliability of the system-level solutions depends on the quality of the surrogate models.
More advanced models can both improve accuracy and widen the space of searchable designs.
Some blocks also admit a restricted set of design choices.
The chip-design block currently assumes
a single configuration of a spatial-array accelerator.
Richer sets of configurations also require a suitable evaluator and surrogate model
behind the same interface, leaving the rest of the co-design
diagram unchanged.

\paragraph{More sophisticated interfaces}
We adopted discretely ordered network structures as an interface between network selection and hardware mapping, which impacts the efficiency of co-design algorithms.
Advances in algebraic deep learning architectures \cite{abbottNeuralCircuitDiagrams2024,abbottWeavesWiresMorphisms2026} provide a better language for hardware-aware network optimization, which also hints at a better interface with proper partial order between those blocks.
A similar drawback appears on process nodes between the migration and fabrication blocks.

\paragraph{Online evaluation and advanced commercial solvers}
Surrogate models in this work are fitted with offline-collected datasets.
They could be replaced by or combined with adaptive sampling strategies that evaluate candidates online based on accumulated evidence.
Those online evaluations can be calls to commercial EDA sign-off tools, which ground the Pareto-optimal solutions in sign-off-accurate estimates of power, performance, and area.
Compared to off-line fitted models, online evaluation promises better sample-efficiency and addresses the evaluation-cost challenge.
The compositional online learning framework in \cite{alharbi2026compositional} provides a principled basis for this extension in co-design theory, with demonstrated sample-efficiency gains over uniform sampling and Bayesian optimization.

\paragraph{Hierarchical system co-design}
Individual \glspl{abk:mdpi} are themselves valid components in a larger
co-design diagram.
The \glsnnp co-design problem studied here could become a sub-block in a
higher-level diagram encompassing PCB integration, thermal constraints, and
system-level reliability requirements, following the same composition rules
applied here to training, mapping, and fabrication.
This compositionality is what distinguishes the monotone co-design framework
from purpose-built joint-search methods: the methodology scales upward without
requiring a new formulation at each level of the design hierarchy.

\section*{Acknowledgments}
The authors used Claude (Anthropic) as a writing assistance tool to
improve the readability and clarity of selected passages in the manuscript.
No AI-generated scientific content, figures, or code appear in this work.

\bibliographystyle{IEEEtran}
\bibliography{ref}

\appendices
\section{Distributional Uncertain \texorpdfstring{\glspl{abk:mdpi}}{MDPIs}}
\label{app:stochastic-mdpi}

This appendix relates the budgeted stochastic \glspl{abk:mdpi} of \cref{def:budgeted-stochastic} to the distributional uncertain \glspl{abk:mdpi} of~\cite{huangDistributionalUncertaintyAdaptive2026} and defines the conditional distributions over $\functionality{F}$.

For each implementation~$\implementation{i}$ and outcome~$\omega$,~$\Phi$ induces the sampled deterministic \gls{abk:dp}
\begin{equation}\label{eq:induced-sampled-dp}
    \dprb_{\Phi,\implementation{i},\omega}
    =
    \makeset{
        \maketup{x_{\functionality{F}},x_{\resource{R}}}
        \mid
        \exists x_{\resource{R}}' \preceq x_{\resource{R}}, x_{\functionality{F}} \preceq \Phi(\implementation{i},x_{\resource{R}}',\omega)
    }.
\end{equation}
The set~$\dprb_{\Phi,\implementation{i},\omega}$ is a valid \glsdp because its existential quantifier implements the \emph{best-so-far} policy of \cref{rmk:best_so_far}.
A larger budget can always be reduced to a smaller budget~$x_{\resource{R}}'$ that achieves the same functionality~$x_{\functionality{F}}$.
Deterministic \glspl{abk:dp} can be added without loss of generality.
With a deterministic part~$\implementation{i} \mapsto \dprb_{\implementation{i}}$, the distributional uncertain \gls{abk:mdpi} is~$\implementation{i} \mapsto \maketup{\outcomeDist, \omega \mapsto \dprb_{\implementation{i}} \cap \dprb_{\Phi, \implementation{i}, \omega}}$.
Its success probability is $\sup_{x_{\resource{R}}' \preceq x_{\resource{R}}} p_\Phi(\implementation{i},x_{\resource{R}}';x_{\functionality{F}})$ if~$\maketup{x_{\functionality{F}},x_{\resource{R}}} \in \dprb_{\implementation{i}}$ and~$0$ otherwise, and its \gls{abk:mdpi} with confidence is \cref{eq:confidence-augmented-mdpi} restricted to the pairs~$\maketup{x_{\functionality{F}},x_{\resource{R}}} \in \dprb_\implementation{i}$.

    The conditional distribution~$\inducedDist{\cdot}{\implementation{i},x_{\resource{R}}}$ of \cref{sec:surrogate-model} is the pushforward of the distribution~$\outcomeDist$ through the map~$\omega \mapsto \Phi(\implementation{i},x_{\resource{R}},\omega)$:
    \begin{equation}\label{eq:induced-distribution}
        \inducedDist{S \subseteq \functionality{F}}{\implementation{i},x_{\resource{R}}}
        =
        \outcomeDist\bigl(\makeset{\omega \in \Omega \mid \Phi(\implementation{i},x_{\resource{R}},\omega) \in S}\bigr)
    \end{equation}
    for every measurable subset~$S \subseteq \functionality{F}$.
    The pushforward defines a map from~$\implementation{I} \times \resource{R}$ to the set~$\DistOf(\functionality{F})$ of distributions over~$\functionality{F}$.

\section{Surrogate Model Provides Lower Bound on the Success Probability}
\label{app:surrogate-proof}

\begin{theorem}\label{thm:lower-bound-by-surrogate-model}
    For a budgeted stochastic \gls{abk:mdpi} with map~$\Phi$, the confidence-augmented \gls{abk:mdpi} defined by \cref{eq:confidence-augmented-mdpi} provides a lower bound on the probability of satisfying the functionality requirement~$x_{\functionality{F}}$ under implementation~$\implementation{i}$ and resource budget~$x_{\resource{R}}$:
    \begin{equation*}
        \sup_{x_{\resource{R}}' \preceq x_{\resource{R}}} p_\Phi(\implementation{i},x_{\resource{R}}';x_{\functionality{F}})
        \leq
        \outcomeDist\left(
            \makeset{\omega \in \Omega \mid \maketup{x_{\functionality{F}},x_{\resource{R}}} \in \dprb_{\Phi,\implementation{i},\omega}}    
        \right).
    \end{equation*}
    If the map $\Phi(\implementation{i},\cdot,\omega)$ is further monotone for every~$\implementation{i}\in\implementation{I}$ and~$\omega\in\Omega$, the inequality becomes an equality.
\end{theorem}
\begin{proof}
    Fix~$\implementation{i}$, $x_{\resource{R}}$, and~$x_{\functionality{F}}$.
    Let $A_{x_{\resource{R}}'} := \makeset{\omega \in \Omega \mid x_{\functionality{F}} \preceq \Phi(\implementation{i},x_{\resource{R}}',\omega)}$ for every~$x_{\resource{R}}' \in \resource{R}$.
    By \cref{eq:success-probability-general,eq:induced-distribution},~$p_\Phi(\implementation{i},x_{\resource{R}}';x_{\functionality{F}}) = \outcomeDist(A_{x_{\resource{R}}'})$, and by \cref{eq:induced-sampled-dp} the right-hand side of the claim is~$\outcomeDist(B)$ with~$B := \bigcup_{x_{\resource{R}}' \preceq x_{\resource{R}}} A_{x_{\resource{R}}'}$.
    We assume that~$B$ is measurable, which holds for every block of this paper: either~$\resource{R}$ is discrete, so the union is finite, or the down-set of~$x_{\resource{R}}$ is compact, the orders are closed, and~$\Phi(\implementation{i},\cdot,\cdot)$ is jointly continuous, so~$B$ is the projection of a closed set along a compact factor and hence closed.
    Since~$A_{x_{\resource{R}}'} \subseteq B$ for every~$x_{\resource{R}}' \preceq x_{\resource{R}}$, each~$p_\Phi(\implementation{i},x_{\resource{R}}';x_{\functionality{F}})$ is at most~$\outcomeDist(B)$, and taking the supremum gives the inequality.
    If~$\Phi(\implementation{i},\cdot,\omega)$ is monotone for every~$\omega$, then~$x_{\resource{R}}' \preceq x_{\resource{R}}$ implies~$A_{x_{\resource{R}}'} \subseteq A_{x_{\resource{R}}}$ by transitivity, so~$B = A_{x_{\resource{R}}}$ and the supremum is attained at~$x_{\resource{R}}' = x_{\resource{R}}$ with value~$\outcomeDist(A_{x_{\resource{R}}}) = \outcomeDist(B)$.
\end{proof}

\section{\texorpdfstring{Details of the Chip Design Pipeline}{Details of the Chip Design Pipeline}}
\label{app:chip-design}
\paragraph{Single-layer LEA distribution at checkpoints}\leavevmode %
The modeling metric satisfies $[\mathtt{X}_t^{(r)}]_\implementation{o}^{-1} \in [h_\implementation{o}, \infty)$, where $h_\implementation{o}$, the minimum of $[\mathtt{X}_t^{(r)}]_\implementation{o}^{-1}$ over all runs $r$ and epochs $t$, is estimated from offline profiling data.
In the mixture of \cref{eq:hw-mixture-pdf}, the peak and tail components are the left-truncated normal distributions $f_1 = f_{\text{TN}}(\cdot\,; h_o, \sigma_1, h_o)$ and $f_3 = f_{\text{TN}}(\cdot\,; \mu_3, \sigma_3, h_o)$, with
\begin{equation}
    f_{\text{TN}}([\mathtt{x}]_o^{-1}; \mu, \sigma, h_o) = \frac{f_{\mathcal{N}}([\mathtt{x}]_o^{-1}; \mu, \sigma^2)}{1 - F_{\mathcal{N}}(h_o; \mu, \sigma^2)} \mathbf{1}([\mathtt{x}]_o^{-1} \ge h_o),
    \label{eq:hw-truncnorm-new}
\end{equation}
and the middle component is the shifted Gamma distribution with shape $k$ and scale $\theta$,
\begin{equation}
    f_2([\mathtt{x}]_o^{-1}) = \frac{([\mathtt{x}]_o^{-1}-h_o)^{k-1}\,
    e^{-([\mathtt{x}]_o^{-1}-h_o)/\theta}}{\Gamma(k)\,\theta^k}
    \,\mathbf{1}([\mathtt{x}]_o^{-1} \ge h_o),
    \label{eq:hw-shifted-gamma}
\end{equation}
where $\Gamma(\cdot)$ is the gamma function.
The parameter vector for preference $\implementation{o}$ at checkpoint $t_i$, $\boldsymbol{\vartheta}_\implementation{o}(t_i) = (w_1,\sigma_1,\,w_2,k,\theta,w_3,\mu_3,\sigma_3)$, is estimated by maximum likelihood from $\makeset{[\mathtt{X}_{t_i}^{(r)}]_\implementation{o}}_{r=1}^R$.

\textit{Recovering the LEA tuple of a single layer}\quad
Since the mixture characterizes only the prioritized metric, we recover the complete \glslea tuple approximately on the one-dimensional support of $\mathcal{S}_t(\mathrm{layer}, \implementation{\chipconfig},  \implementation{o})$.

\begin{figure}[t]
    \centering
    \includegraphics[width=0.2\textwidth]{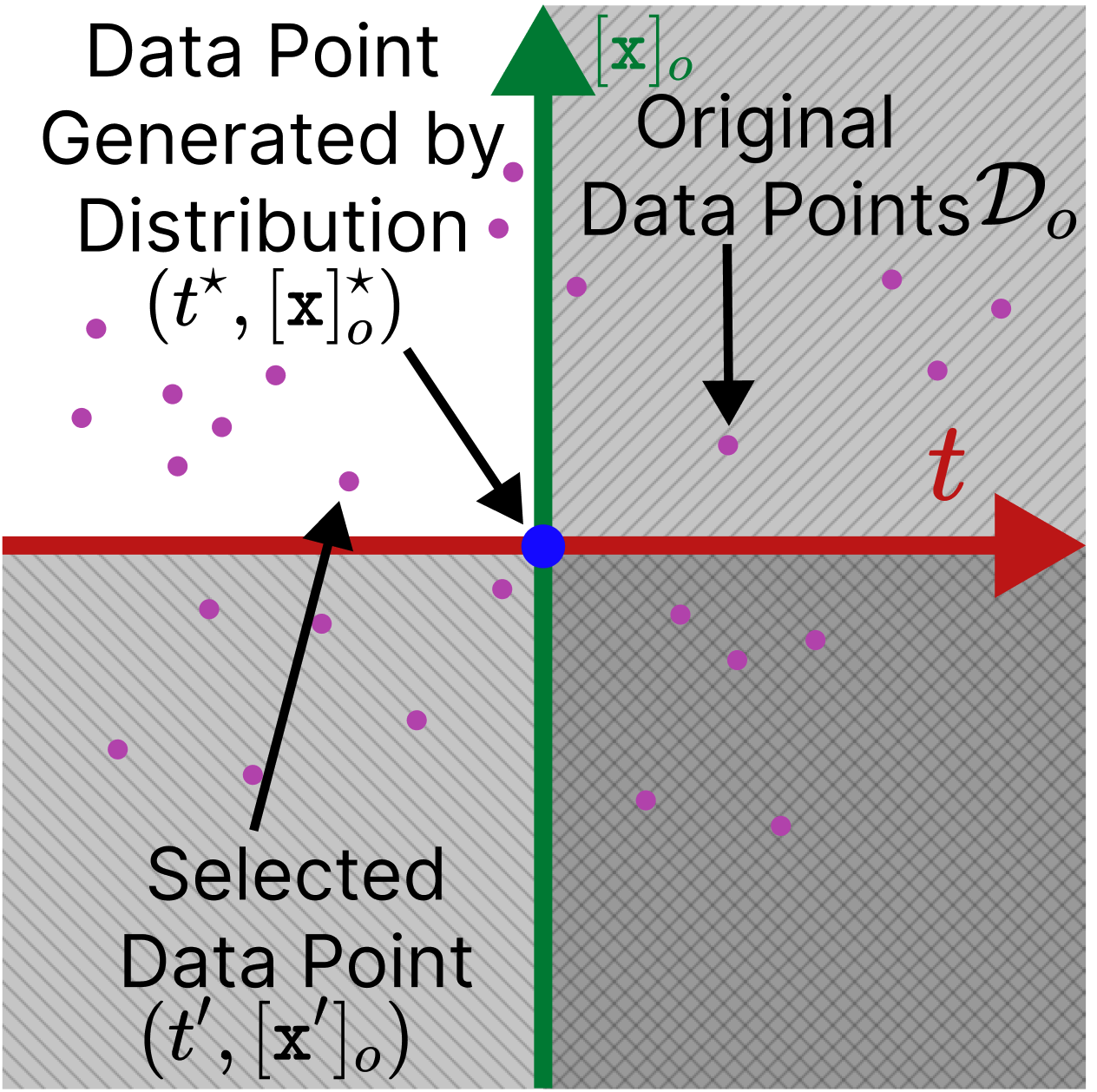}
    \caption{Nearest-neighbor in the $(t, [\mathtt{x}]_o)$ space. Only data points in the second quadrant are valid candidates. Data points in shaded regions are invalid.}
    \label{fig:hw-valid-region}
\end{figure}
For each preference $\implementation{o}$, let $\mathcal{D}_\implementation{o} = \makeset{\left(t_i,\, \mathtt{X}_{t_i}^{(r)}\right)}_{i, r}$ be the profiling data of epoch numbers and observed \glslea tuples.
For a budget~$t^\star$ and a value~$[\mathtt{x}]_o^\star$ of the prioritized metric sampled from \cref{eq:hw-mixture-pdf}, we take the nearest valid observation in~$\mathcal{D}_{\implementation{o}}$, as shown in \cref{fig:hw-valid-region}:
\begin{equation}
    (\hat{t}, \text{}\hat{\mathtt{x}}) = \operatorname*{arg\,min}_{
    \substack{
    (t',\, \mathtt{x}') \in \mathcal{D}_\implementation{o} \\ 
    t' \le t^\star,\; [\mathtt{x}']_o \geq [\mathtt{x}]_o^\star
    }
    }
    \sqrt{(t' - t^\star)^2 + \left([\mathtt{x}']_o \mathbin{\text{}-} [\mathtt{x}]_o^\star\right)^2}.
    \label{eq:hw-nn-lookup}
\end{equation}
The chosen observation uses at most the requested budget and is at least as good in the preferred metric.
The remaining \glslea coordinates are read from  \(\hat{\mathtt{x}}\), preserving the empirical coupling among area, energy, and latency.

\paragraph{Parameter regression over solver budget} To estimate the parameters at epochs that are not profiled, we regress each component $\vartheta(t)$ of $\boldsymbol{\vartheta}_\implementation{o}(t)$ on~$t$.
From the maximum-likelihood estimates $\{(t_i,\, \vartheta_i)\}_{i=1}^{T}$ at the $T$ checkpoints, we select the curve with the smallest \glsrmse in the family $\mathcal{G}$ of linear, power-law, sigmoid, and exponential functions,
\begin{equation}
    g^\star = \operatorname*{argmin}_{g \,\in\, \mathcal{G}}\;
    \sqrt{\frac{1}{T}\sum_{i=1}^{T} \bigl(g(t_i) - \vartheta_i\bigr)^2}.
    \label{eq:hw-param-regression}
\end{equation}
enforcing positivity of the scale and shape parameters and projecting the mixture weights onto the probability simplex.

\end{document}